 \documentclass[accepted]{uai2023} % after acceptance, for a revised
\usepackage[american]{babel}
\usepackage{natbib} % has a nice set of citation styles and commands
\usepackage{mathtools} % amsmath with fixes and additions
\usepackage{booktabs} % commands to create good-looking tables
\usepackage{tikz} % nice language for creating drawings and diagrams

\usepackage{algorithm}
\usepackage{algorithmic}
\usepackage{amsthm}
\usepackage{subfigure}
\usepackage{epsfig}
\usepackage{graphicx}
\usepackage{url}
\usepackage{multirow}
\usepackage{amssymb}
\usepackage{amsthm}
\usepackage{mathrsfs}
\usepackage{epstopdf}
\usepackage{multicol}
\usepackage{amsmath}

\newtheorem{theorem}{\textbf{Theorem}}
\newtheorem{assumption}{\textbf{Assumption}}

\newtheorem{corollary}{\textbf{Corollary}}
\newtheorem{remark}{\textbf{Remark}}
\newtheorem{definition}{\textbf{Definition}}
\newtheorem{proposition}{\textbf{Proposition}}

\title{Can Decentralized Stochastic Minimax Optimization Algorithms Converge Linearly for Finite-Sum Nonconvex-Nonconcave Problems?}
\author[1]{Yihan Zhang}
\author[2]{Wenhao Jiang}
\author[3]{Feng Zheng}
\author[1]{Chiu C. Tan}
\author[1]{Xinghua Shi}
\author[1]{Hongchang Gao}

\affil[1]{%
Temple University 
}
\affil[2]{%
   Tencent
}
\affil[3]{%
    Southern University of Science and Technology
  }
  
\begin{document}
\maketitle

\begin{abstract}
Decentralized minimax optimization has been actively studied in the past few years due to its application in a wide range of machine learning models. However, the current theoretical understanding of its convergence rate is far from satisfactory, since existing works only focus on the nonconvex-strongly-concave problem. This motivates us to study decentralized minimax optimization algorithms for the nonconvex-nonconcave problem. To this end, we develop two novel decentralized stochastic variance-reduced gradient descent ascent algorithms for the finite-sum nonconvex-nonconcave problem that satisfies the Polyak-Łojasiewicz (PL) condition. In particular, our theoretical analyses demonstrate how to conduct local update and perform communication to achieve the linear convergence rate. To the best of our knowledge, this is the first work achieving linear convergence rates for decentralized nonconvex-nonconcave problems. Finally, we verify the performance of our algorithms on both synthetic and real-world datasets. The experimental results confirm the efficacy of our algorithms. 
\end{abstract}

\section{Introduction}
A wide range of  machine learning models can be formulated as a minimax optimization problem, such as adversarially robust machine learning models \cite{goodfellow2014generative,goodfellow2014explaining,madry2017towards}, distributionally robust machine learning models \cite{lin2020gradient,luo2020stochastic}, AUC maximization models \cite{ying2016stochastic,liu2019stochastic}, just name a few. To facilitate such kinds of  machine learning models for distributed data, this paper studies the decentralized  finite-sum minimax problem, which is defined as follows:
\begin{equation} \label{eq_loss}
	\begin{aligned}
				& \min_{{x}\in \mathbb{R}^{d_x}} \max_{{y}\in \mathbb{R}^{d_y}} f({x}, {y}) \triangleq \frac{1}{K}\sum_{k=1}^{K}\Big(\frac{1}{n}\sum_{i=1}^{n} f_i^{(k)}({x}, {y})\Big) \ ,
	\end{aligned}
\end{equation}
where $k\in \{1, 2, \cdots, K\}$ is the index of worker, $f^{(k)}({x}, {y})\triangleq\frac{1}{n}\sum_{i=1}^{n} f_i^{(k)}({x}, {y})$ denotes the loss function on the $k$-th worker, $f_i^{(k)}({x}, {y})$ represents the loss function of the $i$-th sample on the $k$-th worker. Here, all workers compose a communication graph and its adjacency matrix is denoted by $W=[w_{ij}]\in\mathbb{R}_{+}^{K\times K}$, where $w_{ij}>0$ denotes worker $i$
 and $j$ are connected, otherwise $w_{ij}=0$. Based on such a communication graph, the worker conducts peer-to-peer communication.

 In this paper, we consider the decentralized nonconvex-nonconcave minimax problem that satisfies the PL condition, which is still unexplored and more challenging than the nonconvex-strongly-concave minimax problem. In fact, such a condition is satisfied in many applications, such as overparameterized neural networks \cite{liu2022loss}, robust phase retrieval model \cite{sun2018geometric}, deep AUC maximization model \cite{liu2019stochastic}, just name a few. Until very recently, the convergence of minimax optimization algorithms under the PL condition has been studied by \cite{yang2020global,chenfaster} under the single-machine setting. In particular, \cite{yang2020global} investigated the   alternating  gradient descent ascent (AGDA) algorithm, which demonstrated that the variance-reduced variant enjoys  the linear convergence rate.  \cite{chenfaster} shows that the variance-reduced stochastic gradient descent ascent (SGDA) algorithms can also achieve the linear convergence rate even though they do not take the alternating update rule. 
 
 Obviously, the aforementioned minimax optimization algorithms show promising performance under the single-machine setting. Then, a natural question arises: \textit{Can decentralized stochastic minimax optimization algorithms converge linearly for  nonconvex-nonconcave problems that satisfy the PL condition?}

In fact, the decentralized setting brings unique challenges to investigate the convergence behavior.  Specifically, each worker under the decentralized setting updates its local model parameters with its own gradient. Additionally, due to the decentralized communication strategy, there does not exist global communication. This results in large consensus errors between local model parameters and global model parameters, which has an adverse effect on the convergence rate. Currently,  it is unclear whether the consensus error will make the linear convergence rate unachievable. Then, it is necessary to study how the consensus error affect the convergence rate and design new algorithms to alleviate the large consensus error issue. 

% it is still unclear how the consensus error affects the linear convergence rate. Especially,

% Therefore, it is important to study how the consensus error affects the linear convergence rate. 
% \vspace{-5pt}
\subsection{Contributions}
% \vspace{-5pt}
To address the aforementioned questions, we develop two decentralized stochastic variance-reduced gradient descent ascent (DSVRGDA) algorithms for the finite-sum minimax optimization problem that satisfies the PL condition. Generally speaking, our study discovered that the traditional decentralized communication strategy cannot make DSVRGDA achieve the linear convergence rate due to the large consensus error. Therefore, to mitigate the large consensus error, we compensate the traditional decentralized communication strategy with the stage-wise multi-round communication. As such, the consensus error can be reset to zero stage-wisely and then our algorithms are able to achieve the linear convergence rate. 
In summary, we made the following contributions in this paper. 
\begin{itemize}
    \item We develop two decentralized stochastic variance-reduced gradient descent ascent algorithms based on the gradient-tracking communication strategy and stage-wise multi-round communication strategy. As far as we know, this is the first time showing how to conduct local update and perform communication for the aforementioned minimax problem.
    \item We establish the convergence rate of our two decentralized algorithms for the finite-sum minimax optimization problem that satisfies the PL condition. They both can achieve the linear convergence rate. To the best of our knowledge, this is the first work achieving linear convergence rate for such a kind of decentralized minimax  problem.
    % \item As a byproduct, we established the convergence rate of our algorithms for  the finite-sum minimax optimization problem that satisfies the nonconvex-PL condition, which can match the algorithm designed for nonconvex-strongly-concave problems. As far as we know, this is also the first work achieving such a kind of convergence rate for decentralized nonconvex-PL minimax optimization problems. 
    \item We evaluate our two algorithms on both synthetic and real-world datasets. The extensive experimental results confirm the effectiveness of our  two algorithms.
\end{itemize}

% \vspace{-10pt}
 \section{Related Works}
 % \vspace{-10pt}
In this section, we briefly introduce the convergence rate of existing minimax optimization and decentralized optimization problems. 

 \subsection{Minimax Optimization}
 Since many emerging machine learning models, e.g., adversarial generative networks \cite{goodfellow2014explaining}, can be formulated as a minimax optimization problem, how to efficiently optimize this kind of problem has been extensively studied in recent years. In particular, 
under the single-machine setting, \cite{lin2020gradient} established the convergence rate of stochastic gradient descent ascent (SGDA) for nonconvex-strongly-concave problems, i.e., $O(1/\epsilon^4)$ to achieve the $\epsilon$-accuracy solution. \cite{luo2020stochastic} improved it to $O(n+n^{1/2}/\epsilon^2)$ for finite-sum problems with the SPIDER \cite{fang2018spider,nguyen2017sarah} gradient estimator. \cite{huang2020accelerated} leveraged the STORM \cite{cutkosky2019momentum} gradient estimator to improve the convergence rate to $O(1/\epsilon^4)$ for stochastic problems. 

When the objective function satisfies the PL condition, \cite{yang2020global} investigated the convergence rate of the alternating gradient descent ascent (AGDA) algorithm, i.e., two variables should be updated sequentially. In particular, \cite{yang2020global} showed that AGDA can achieve the linear convergence rate when using full gradient, but it fails to achieve that when employing stochastic gradient. Then, \cite{yang2020global} exploits the variance-reduced gradient estimator \cite{johnson2013accelerating} to make the algorithm achieve linear convergence rate. Recently, \cite{chenfaster} studied the convergence rate of SGDA for the mininax optimization problem that satisfies PL conditions. In particular, \cite{chenfaster} developed two variance-reduced SGDA algorithms based on \cite{johnson2013accelerating} and \cite{fang2018spider,nguyen2017sarah} respectively, which demonstrated a linear convergence rate for finite-sum problems. All aforementioned algorithms only concentrate on the single-machine setting so that they cannot be utilized to optimize Eq.~(\ref{eq_loss}).

 \subsection{Decentralized Optimization}
Decentralized optimization has been widely leveraged to train machine learning models for distributed data. Currently, numerous algorithms have been proposed for the minimization problem. For example, \cite{lian2017can} developed the decentralized stochastic gradient descent (DSGD) algorithm based on the gossip communication strategy and demonstrated that the communication topology only affects the high-order term of the convergence rate for nonconvex problems. \cite{lu2019gnsd} proposed a DSGD algorithm based on the gradient-tracking communication scheme. After than, \cite{sun2020improving,xin2020near} improved the convergence rate of DSGD with variance-reduced gradient estimators \cite{fang2018spider,nguyen2017sarah,cutkosky2019momentum} for finite-sum and stochastic problems. Since these algorithms are designed for minimization problem, they cannot be used to optimize Eq.~(\ref{eq_loss}).

As for decentralized minimax optimization, \cite{xian2021faster} proposed a decentralized stochastic gradient descent ascent with the STORM \cite{cutkosky2019momentum} gradient estimator algorithm and established its convergence rate for stochastic nonconvex-strongly-concave problems. \cite{zhang2021taming} employed the SPIDER \cite{fang2018spider,nguyen2017sarah} gradient estimator for decentralized SGDA and provided its convergence rate for finite-sum problems. Later, \cite{gao2022decentralized} resorted to the ZeroSARAH \cite{li2021zerosarah} gradient estimator and then improved the convergence rate of \cite{zhang2021taming}. More recently, \cite{chen2022simple} further improved the communication complexity of \cite{zhang2021taming,gao2022decentralized} with a multi-step communication strategy, but suffers from a worse sample complexity than \cite{gao2022decentralized}. All these theoretical convergence results only hold for nonconvex-strongly-concave problems. It is still unclear how decentralized minimax optimization algorithms converge under the PL condition.

 \section{Preliminaries}
In this section, we provide some fundamental definitions and assumptions for studying the convergence rate of our algorithms.

\begin{definition}
A function $f(x): \mathbb{R}^d\rightarrow \mathbb{R}$ satisfies the $\mu$-PL (Polyak-Łojasiewicz) condition if there exists $\mu>0$ such that
\begin{equation}
    2\mu(f(x) - f(x_*))\leq \|\nabla f(x)\|^2, \ \forall x \in \mathbb{R}^d \  , 
\end{equation}
where $x_*=\arg\min_{x' \in \mathbb{R}^d} f(x')$.
\end{definition}
It is worth noting that the PL condition is  weaker than strong convexity. A strongly convex function satisfies the PL condition, but not vice versa. For instance, a nonconvex function, e.g., the overparameterized neural network \cite{liu2022loss}, can also satisfy the PL condition.

Based on this definition, we introduce the following assumptions for the loss function in Eq.~(\ref{eq_loss}). 
\begin{assumption}\label{assumption_pl}
The function $f(\cdot, \tilde{y})$ satisfies the $\mu$-PL condition for any fixed $\tilde{y}\in \mathbb{R}^{d_y}$, i.e., for $\forall x \in \mathbb{R}^{d_x}$, there exists $\mu>0$ such that
\begin{equation}
    \begin{aligned}
        & \quad 2\mu(f(x, \tilde{y}) - \min_{x' \in \mathbb{R}^{d_x}} f(x', \tilde{y}))  \leq \|\nabla_{x} f(x, \tilde{y})\|^2.   
    \end{aligned}
\end{equation}
Similarly, $-f(\tilde{x}, \cdot)$ satisfies the $\mu$-PL condition for any fixed $ \tilde{x}\in \mathbb{R}^{d_x}$, i.e., for $\forall y \in \mathbb{R}^{d_y}$, there exists $\mu>0$ such that
\begin{equation}
    \begin{aligned}
        &  2\mu(-f(\tilde{x}, y) + \max_{y' \in \mathbb{R}^{d_y}} f(\tilde{x}, y')) \leq \|\nabla_{y} f(\tilde{x}, y)\|^2  .  \\
    \end{aligned}
\end{equation}
\end{assumption}

\begin{assumption}\label{assumption_smooth}
    For $\forall k \in\{1, 2, \cdots, K\}$ and $\forall i \in \{1, 2, \cdots, n\}$, $f_i^{(k)}(\cdot, \cdot)$ is $L$-smooth, i.e.,
    \begin{equation}
    \begin{aligned}
        &  \quad \|\nabla f_i^{(k)}(x_1, y_1) - \nabla f_i^{(k)}(x_2, y_2)\|^2 \\
        & \leq L^2\|x_1-x_2\|^2 + L^2\|y_1 -y_2\|^2 \  ,
    \end{aligned}
    \end{equation}
    where $L>0$ is a constant value, 
\end{assumption}
Based on Assumption~\ref{assumption_pl}-\ref{assumption_smooth}, we denote the condition number as $\kappa=L/\mu$. 

% Under the decentralized setting, the workers compose a communication network, which can be represented as a graph $\mathcal{G}=\{\mathcal{P}, W\}$, where $\mathcal{P}$ denotes the worker set $\mathcal{P}=\{1, 2, \cdots, K\}$, $W=[w_{ij}]\in \mathbb{R}_{+}^{K\times K}$ denotes the adjacency matrix. Moreover, 
Under the decentralized setting, we have the following assumptions for the adjacency matrix, which are commonly used in existing decentralized optimization literature \cite{xian2021faster,zhang2021taming,gao2022decentralized}
\begin{assumption} \label{assumption_graph}
    % 1) If worker $i$ and $j$ are connected, $w_{ij}>0$. Otherwise, $w_{ij}=0$. 
    1) $W$ is doubly stochastic and symmetric.  2) Its eigenvalues satisfy $|\lambda_K|\leq |\lambda_{K-1}|\leq \cdots \leq |\lambda_2|<|\lambda_1|=1$.
\end{assumption}
Based on Assumption~\ref{assumption_graph}, we denote $\lambda=|\lambda_2|$ so that  the spectral gap can be represented as $1-\lambda$. Moreover, we denote the diameter of the communication graph as $D$.

Under these assumptions, the goal of this work is to find the $\epsilon$-saddle point of Eq.~(\ref{eq_loss}), i.e., a solution $x$ satisfies $g(x)-g(x^*)\leq \epsilon$ where $g(x)\triangleq \max_{y\in\mathbb{R}^{d_y}} f(x, y)$.  Here, we assume the saddle point does exist. In particular, we have the following assumption.
\begin{assumption} \cite{yang2020global,chenfaster}
    The function $f(x, y)$ has at least one saddle point $(x^*, y^*)$. Moreover, for any fixed $\tilde{y}\in \mathbb{R}^{d_y}$, the solution set of the subproblem $\min_{x\in\mathbb{R}^{d_x}} f(x, \tilde{y})$ is not empty and its optimal value is finite. This is also true for the subproblem  $\min_{y\in\mathbb{R}^{d_y}} f(\tilde{x}, y)$ when fixing $\tilde{x}\in \mathbb{R}^{d_x}$.
\end{assumption}
Furthermore, from Lemma A.5 in \cite{nouiehed2019solving}, we can know that $g(x)$ is $L_g$-smooth where $L_g=2L^2/\mu$. 

% Throughout this paper, we use $a_{s,r}^{(k)}$ denotes the variable $a$ in the $r$-th iteration of $s$-th stage on the $k$-th worker. 

\begin{algorithm}[h]
	\caption{DSVRGDA-P}
	\label{dpage}
	\begin{algorithmic}[1]
		\REQUIRE $\tilde{x}_{0}^{(k)}={x}_{0}$,  $\tilde{y}_{0}^{(k)}={y}_{0}$,   $\eta_x>0$, $\eta_y>0$.   
		\FOR{$s=0,\cdots, S-1$, each worker $k$} 
		\STATE $x_{s, 0}^{(k)}= \tilde{x}_{s}^{(k)}$, $y_{s, 0}^{(k)}= \tilde{y}_{s}^{(k)}$ ,
		\FOR{$r=0,\cdots, R-1$, each worker $k$} 
		\IF {$r==0$}
		\STATE $\phi_{s,r}^{(k)}=\nabla_x f^{(k)}(x_{s,r}^{(k)}, y_{s,r}^{(k)}) $ ,\\
		 $\psi_{s,r}^{(k)}=\nabla_y f^{(k)}(x_{s,r}^{(k)}, y_{s,r}^{(k)}) $ ,\\
		\ELSE 
        \STATE Compute the gradient estimator $\phi_{s,r}^{(k)}$ and $\psi_{s,r}^{(k)}$ as Eq.~(\ref{eq_page_x}) and Eq.~(\ref{eq_page_y}),
		\ENDIF
		\IF {$r==0$}
		\STATE $u_{s, r}^{(k)} =\phi_{s,r}^{(k)}$, 
		 $v_{s, r}^{(k)} =  \psi_{s,r}^{(k)}$,
		\ELSE 
		\STATE $u_{s, r}^{(k)} = \sum_{j\in \mathcal{N}_{k}}w_{kj}u_{s, r-1}^{(k)}  - \phi_{s,r-1}^{(k)} + \phi_{s,r}^{(k)}$\ , \\
		 $v_{s, r}^{(k)} = \sum_{j\in \mathcal{N}_{k}}w_{kj}v_{s, r-1}^{(k)}  - \psi_{s,r-1}^{(k)} + \psi_{s,r}^{(k)}$\ ,
		\ENDIF 
		\STATE ${x}_{s, r+1}^{(k)}=\sum_{j\in \mathcal{N}_{k}}w_{kj}{x}_{s, r}^{(j)} - \eta_x u_{s, r}^{(k)}$\ , \\
		 ${y}_{s, r+1}^{(k)}=\sum_{j\in \mathcal{N}_{k}}w_{kj}{y}_{s, r}^{(j)} + \eta_y v_{s, r}^{(k)}$\ ,
		\ENDFOR
		
		\STATE Randomly choose $(\tilde{x}_{s+1}^{(k)}, \tilde{y}_{s+1}^{(k)})$  from $\{(\tilde{x}_{s, r}^{(k)}, \tilde{y}_{s, r}^{(k)})\}_{r=0}^{R-1}$ with the same random seed  for all workers, \\
		\STATE  Conduct $D$-round communication such that \\
  $\tilde{x}_{s+1}^{(k)}=\frac{1}{K}\sum_{k'=1}^{K}\tilde{x}_{s+1}^{(k')}$ , 
  $\tilde{y}_{s+1}^{(k)}=\frac{1}{K}\sum_{k'=1}^{K}\tilde{y}_{s+1}^{(k')}$ , 
		\ENDFOR
	\end{algorithmic}
\end{algorithm}

\section{Decentralized Minimax Optimization Algorithms}
In this section, we present the detail of our  decentralized optimization algorithms to show how to do local update and perform communication.

\subsection{DSVRGDA-P}
In Algorithm~\ref{dpage}, we proposed a novel decentralized stochastic variance-reduced gradient descent ascent algorithm DSVRGDA-P based on the PAGE gradient estimator \cite{li2021page}. Generally speaking, this is a stage-wise algorithm. In detail, in each stage $s$, each worker $k$ is initialized with the same model parameters, i.e., $x_{s, 0}^{(k)}= \tilde{x}_{s}^{(k)}$, $y_{s, 0}^{(k)}= \tilde{y}_{s}^{(k)}$. In fact, this step is critical to achieve linear convergence rate, which will be discussed in details in the next section. Then, in each inner iteration $r$, each worker $k$ computes the  gradient estimator for the variable $x$ as follows:
\begin{equation} \label{eq_page_x}
    \begin{aligned}
     &   \phi_{s,r}^{(k)}= \begin{cases}
\nabla_x f^{(k)}(x_{s,r}^{(k)}, y_{s,r}^{(k)}),  & \text{with probability $p$} \\
 \phi_{s,r-1}^{(k)} + \Delta_{ s, r}^{(k)}(x), & \text{with probability $1-p$} \\
\end{cases} \ , 
    \end{aligned}
\end{equation}
where $\Delta_{s, r}^{(k)}(x)=\frac{1}{b}\sum_{i\in \mathcal{B}_{s,r}^{(k)}}( \nabla_x f_i^{(k)}(x_{s,r}^{(k)}, y_{s,r}^{(k)})  - \nabla_x f_i^{(k)}(x_{s,r-1}^{(k)}, y_{s,r-1}^{(k)}))$, $\mathcal{B}_{s,r}^{(k)}$ is the randomly selected mini-batch with the batch size $b$. Similarly, each worker $k$ computes the gradient estimator for the variable $y$ as follows:
\begin{equation}\label{eq_page_y}
    \begin{aligned}
     &   \psi_{s,r}^{(k)}= \begin{cases}
\nabla_y f^{(k)}(x_{s,r}^{(k)}, y_{s,r}^{(k)}),  & \text{with probability $p$} \\
 \psi_{s,r-1}^{(k)} + \Delta_{s, r}^{(k)}(y), & \text{with probability $1-p$} \\
\end{cases} \ , 
    \end{aligned}
\end{equation}
where $\Delta_{s, r}^{(k)}(y)=\frac{1}{b}\sum_{i\in \mathcal{B}_{s,r}^{(k)}}( \nabla_y f_i^{(k)}(x_{s,r}^{(k)}, y_{s,r}^{(k)})  - \nabla_y f_i^{(k)}(x_{s,r-1}^{(k)}, y_{s,r-1}^{(k)}))$. 

After that, each worker $k$ leverages the gradient-tracking communication strategy to compute  $u_{s, r}^{(k)}$ and $v_{s, r}^{(k)}$ as shown in Step 12 of Algorithm~\ref{dpage}, where $\mathcal{N}_k$ includes the neighboring workers of the $k$-th worker and itself. Then, based on these gradients, each worker $k$ updates it local model parameters ${x}_{s, r+1}^{(k)}$ and ${y}_{s, r+1}^{(k)}$ as shown in Step 14 of Algorithm~\ref{dpage}, where $\eta_x>0$ and $\eta_y>0$ are the learning rate. 

After finishing the inner loop, each worker $k$ randomly selects a solution from $\{(\tilde{x}_{s, r}^{(k)}, \tilde{y}_{s, r}^{(k)})\}_{r=0}^{R-1}$ with the same random seed as $(\tilde{x}_{s+1}^{(k)}, \tilde{y}_{s+1}^{(k)})$. Then, in Step 17, DSVRGDA-P conducts multi-round communication to obtain the global variable. In fact, it can be done in $D$ round  for a graph whose diameter is $D$ \cite{vogels2021relaysum}. For instance, it can be done in $D=K-1$ communication round for a line graph \cite{scaman2017optimal}. \textit{This stage-wisely multi-round communication strategy is critical to achieve the linear convergence rate, which will be demonstrated in next section. }

To the best of our knowledge, this is the first work proposing this kind of communication strategy to guarantee the linear convergence rate. Thus, our algorithmic design is novel. 

\begin{algorithm}[!h]
	\caption{DSVRGDA-Z}
	\label{alg_dsvrgd_z}
	\begin{algorithmic}[1]
		\REQUIRE $\tilde{x}_{0}^{(k)}={x}_{0}$,  $\tilde{y}_{0}^{(k)}={y}_{0}$,  $\eta_x>0$, $\eta_y>0$.  
		%			\REQUIRE ${x}_{0}^{(k)}={x}_{-1}^{(k)}={x}_{0}$, ${y}_{0}^{(k)}={y}_{-1}^{(k)}={y}_{0}$, ${v}_{-1}^{(k)}={a}_{-1}^{(k)}=0$, ${u}_{-1}^{(k)}={b}_{-1}^{(k)}=0$,\\ 
		\FOR{$s=0,\cdots, S-1$, each worker $k$} 
		\STATE $x_{s, 0}^{(k)}= \tilde{x}_{s}^{(k)}$, $y_{s, 0}^{(k)}= \tilde{y}_{s}^{(k)}$, \\
		${g}_{s,-1}^{(k)}(i)=0$, ${h}_{s,-1}^{(k)}(i)=0$ for $i\in \{1,2,\cdots n\}$, 
		\FOR{$r=0,\cdots, R-1$, each worker $k$} 
		\IF {$r==0$}
		\STATE $\phi_{s, r}^{(k)}=\frac{1}{b_0} \sum_{i \in {\mathcal{B}}_{s, 0}}\nabla_{{x}} f^{(k)}_{i}({x}_{s, r}^{(k)}, {y}_{s, r}^{(k)}) $ , \\
		 $\psi_{s, r}^{(k)}=\frac{1}{b_0} \sum_{i \in {\mathcal{B}}_{s, 0}}\nabla_{{y}} f^{(k)}_{i}({x}_{s, r}^{(k)}, {y}_{s, r}^{(k)})$ , \\
		\ELSE

		\STATE Compute the gradient estimator $\phi_{s,r}^{(k)}$ and $\psi_{s,r}^{(k)}$ as Eq.~(\ref{eq_zero_x}) and Eq.~(\ref{eq_zero_y}),

		% $\phi_{s,r}^{(k)}=\frac{1}{s_1} \sum_{i \in \mathcal{S}_{s, r}}\Big(\nabla_{{x}} f^{(k)}_{i}({x}_{s, r}^{(k)}, {y}_{s, r}^{(k)})-\nabla_{{x}} f^{(k)}_{i}({x}_{s, r-1}^{(k)}, {y}_{s, r-1}^{(k)})\Big)+(1-\rho) \phi_{s, r-1}^{(k)}+\rho\Big(\frac{1}{s_1} \sum_{i \in \mathcal{S}_{s, r}}(\nabla_{{x}} f^{(k)}_{i}({x}_{s, r-1}^{(k)}, {y}_{s, r-1}^{(k)})-{g}_{s, r-1}^{(k)}(i))+\frac{1}{n} \sum_{j=1}^{n} {g}_{s, r-1}^{(k)}(j)\Big)$

		% $\psi_{s,r}^{(k)}=\frac{1}{s_1} \sum_{i \in \mathcal{S}_{s, r}}\Big(\nabla_{{y}} f^{(k)}_{i}({x}_{s, r}^{(k)}, {y}_{s, r}^{(k)})-\nabla_{{y}} f^{(k)}_{i}({x}_{s, r-1}^{(k)}, {y}_{s, r-1}^{(k)})\Big)+(1-\rho) \psi_{s,r-1}^{(k)}+\rho\Big(\frac{1}{s_1} \sum_{i \in \mathcal{S}_{s, r}}(\nabla_{{y}} f^{(k)}_{i}({x}_{s, r-1}^{(k)}, {y}_{s, r-1}^{(k)})-{h}_{s, r-1}^{(k)}(i))+\frac{1}{n} \sum_{j=1}^{n} {h}_{s, r-1}^{(k)}(j)\Big)$
		\ENDIF
		
		\IF {$r==0$}
		\STATE $u_{s, r}^{(k)} =\phi_{s,r}^{(k)}$ ,  
		 $v_{s, r}^{(k)} =  \psi_{s,r}^{(k)}$ , 
		\ELSE 
		\STATE $u_{s, r}^{(k)} = \sum_{j\in \mathcal{N}_{k}}w_{kj}u_{s, r-1}^{(k)}  - \phi_{s,r-1}^{(k)}+ \phi_{s,r}^{(k)}$ ,  \\
		 $v_{s, r}^{(k)} = \sum_{j\in \mathcal{N}_{k}}w_{kj}v_{s, r-1}^{(k)}  - \psi_{s,r-1}^{(k)}+ \psi_{s,r}^{(k)}$ , 
		\ENDIF 
		\STATE ${x}_{s, r+1}^{(k)}=\sum_{j\in \mathcal{N}_{k}}w_{kj}{x}_{s, r}^{(j)} - \eta_x u_{s, r}^{(k)} $ ,  \\
		 ${y}_{s, r+1}^{(k)}=\sum_{j\in \mathcal{N}_{k}}w_{kj}{y}_{s, r}^{(j)} - \eta_y v_{s, r}^{(k)} $ , 
		
		\STATE Update ${g}$ and ${h}$:
		
		${g}_{s, r}^{(k)} (i)= \begin{cases}
			\nabla_{{x}} f^{(k)}_{i}({x}_{s,r}^{(k)} , {y}_{s,r}^{(k)} ),  & \text { for } i \in  \mathcal{B}_{s,r}^{(k)}\\
			{g}_{s, r-1}^{(k)} (i) , & \text{otherwise} \\
		\end{cases} $

		${h}_{s, r }^{(k)}(i) = \begin{cases}
			\nabla_{{y}} f^{(k)}_{i}({x}_{s,r}^{(k)} , {y}_{s,r}^{(k)} ),  & \text { for } i \in  \mathcal{B}_{s,r}^{(k)}\\
			{h}_{s, r-1}^{(k)} (i), & \text{otherwise} \\
		\end{cases} $
		
		\ENDFOR
		
		\STATE Randomly choose $(\tilde{x}_{s+1}^{(k)}, \tilde{y}_{s+1}^{(k)})$  from $\{(\tilde{x}_{s, r}^{(k)}, \tilde{y}_{s, r}^{(k)})\}_{r=0}^{R-1}$ with the same random seed  for all workers, \\
  		\STATE  Conduct $D$-round communication such that \\
		  $\tilde{x}_{s+1}^{(k)}=\frac{1}{K}\sum_{k'=1}^{K}\tilde{x}_{s+1}^{(k')}$, 
    $\tilde{y}_{s+1}^{(k)}=\frac{1}{K}\sum_{k'=1}^{K}\tilde{y}_{s+1}^{(k')}$ ,
		\ENDFOR
	\end{algorithmic}
\end{algorithm}

\subsection{DSVRGDA-Z}
From Eq.~(\ref{eq_page_x}) and Eq.~(\ref{eq_page_y}), it can be observed that the full gradient should be computed with the probability $p$. Then, to avoid the computation of the full gradient, we propose a novel stochastic variance-reduced gradient descent ascent algorithm DSVRGDA-Z based on the ZeroSARAH gradient estimator \cite{li2021zerosarah}, which is shown in Algorithm~\ref{alg_dsvrgd_z}. In detail, each worker $k$ computes the gradient estimator $\phi_{s, r}^{(k)}$ as follows:
\begin{equation} \label{eq_zero_x}
    \begin{aligned}
        & \phi_{s,r}^{(k)}=(1-\rho) \phi_{s, r-1}^{(k)}   +\rho\Big(\frac{1}{n} \sum_{j=1}^{n} {g}_{s, r-1}^{(k)}(j) \\
        &  +\frac{1}{b_1} \sum_{i \in \mathcal{B}_{s, r}^{(k)}}(\nabla_{{x}} f^{(k)}_{i}({x}_{s, r-1}^{(k)}, {y}_{s, r-1}^{(k)})-{g}_{s, r-1}^{(k)}(i))  \Big)  \\
        &  +\frac{1}{b_1} \sum_{i \in \mathcal{B}_{s, r}^{(k)}}\Big(\nabla_{{x}} f^{(k)}_{i}({x}_{s, r}^{(k)}, {y}_{s, r}^{(k)})-\nabla_{{x}} f^{(k)}_{i}({x}_{s, r-1}^{(k)}, {y}_{s, r-1}^{(k)})\Big), \\
    \end{aligned}
\end{equation}
where $\rho>0$ is a constant value, $b_1=|\mathcal{B}_{s, r}^{(k)}|$, ${g}_{s, r-1}^{(k)}(i)$ stores the stochastic gradient of the $i$-th sample, which is updated as shown in Step 19 of Algorithm~\ref{alg_dsvrgd_z}. Obviously, when $\rho=0$, this gradient estimator degenerates to SPIDER.  An advantage of $\phi_{s, r}^{(k)}$ is that we don't need to compute the full gradient so that it is more efficient than Algorithm~\ref{dpage}, especially for a deep neural network where computing gradients through backpropagation is time-consuming. Similarly, each worker $k$ computes the gradient estimator $\psi_{s, r}^{(k)}$ as follows:
\begin{equation} \label{eq_zero_y}
    \begin{aligned}
        & \psi_{s,r}^{(k)}=(1-\rho) \psi_{s,r-1}^{(k)} +\rho\Big(\frac{1}{n} \sum_{j=1}^{n} {h}_{s, r-1}^{(k)}(j)\\
        & + \frac{1}{b_1} \sum_{i \in \mathcal{B}_{s, r}^{(k)}}(\nabla_{{y}} f^{(k)}_{i}({x}_{s, r-1}^{(k)}, {y}_{s, r-1}^{(k)})-{h}_{s, r-1}^{(k)}(i))\Big) \\
        & +\frac{1}{b_1} \sum_{i \in \mathcal{B}_{s, r}^{(k)}}\Big(\nabla_{{y}} f^{(k)}_{i}({x}_{s, r}^{(k)}, {y}_{s, r}^{(k)})-\nabla_{{y}} f^{(k)}_{i}({x}_{s, r-1}^{(k)}, {y}_{s, r-1}^{(k)})\Big) \ , \\
    \end{aligned}
\end{equation}
where ${h}_{s, r-1}^{(k)}(i)$ stores the stochastic gradient of the $i$-th sample and updated in Step 15 of Algorithm~\ref{alg_dsvrgd_z}. The communication and local update for gradients and model parameters are the same with Algorithm~\ref{dpage}.

In summary, we develop two novel decentralized optimization algorithms for optimizing Eq.~(\ref{eq_loss}). With this novel design regarding local update and stage-wise communication, our algorithms are able to achieve the linear convergence rate when the objective function satisfies the PL condition, which will be shown in next section. 

\section{Convergence Rate}
In this section, we present the convergence rate of our two algorithms, especially showing why the stage-wise global communication is necessary. 

To investigate the theoretical convergence rate of our algorithms, we employ the following metric, which is also used in existing works \cite{chenfaster,yang2020global}. 
\begin{equation}
    \mathcal{M}_s\triangleq\mathbb{E}[g(\bar{\tilde{x}}_{s})-g(x_*)] + \frac{c_0 \eta_{x}}{\eta_y}\mathbb{E}[g(\bar{\tilde{x}}_{s})-f(\bar{\tilde{x}}_{s}, \bar{\tilde{y}}_{s})]  \ , 
\end{equation}
where $c_0=\frac{16 L^2}{\mu^2}$, $\bar{\tilde{x}}_{s}=\frac{1}{K}\sum_{k=1}^{K}\tilde{x}_{s}^{(k)}$, $\bar{\tilde{y}}_{s}=\frac{1}{K}\sum_{k=1}^{K}\tilde{y}_{s}^{(k)}$.

\subsection{Convergence Rate of Algorithm~\ref{dpage}}
To establish the convergence rate of Algorithm~\ref{dpage}, we propose a novel potential function, which is defined as follows:
\begin{equation} \label{eq_potential_p}
	\begin{aligned}
		& \mathcal{P}_{s, r}  = \mathbb{E}[g(\bar{x}_{s, r})-g(x_*)] + \frac{c_0 \eta_{x}}{\eta_y}\mathbb{E}[g(\bar{x}_{s, r})-f(\bar{x}_{s, r}, \bar{y}_{s, r})] \\
  & \quad + c_1 \frac{1}{K}\mathbb{E}[\|X_{s, r} - \bar{X}_{s, r}\|_F^2]+ c_2  \frac{1}{K}\mathbb{E}[\|Y_{s, r} - \bar{Y}_{s, r}\|_F^2]  \\
		& \quad + c_3  \frac{1}{K}\mathbb{E}[\| U_{s, r} - \bar{U}_{s, r} \|_F^2] + c_4  \frac{1}{K} \mathbb{E}[\| V_{s, r} - \bar{V}_{s, r} \|_F^2] \\
		& \quad + c_5 \frac{1}{K}\sum_{k=1}^{K}\mathbb{E}[\|\phi^{(k)}(x^{(k)}_{s,r}, y^{(k)}_{s,r})  -\nabla_x f^{(k)}(x^{(k)}_{s,r}, y^{(k)}_{s,r}) \|^2]  \\
		& \quad + c_6 \frac{1}{K}\sum_{k=1}^{K}\mathbb{E}[\|\psi^{(k)}(x^{(k)}_{s,r}, y^{(k)}_{s,r})  -\nabla_y f^{(k)}(x^{(k)}_{s,r}, y^{(k)}_{s,r}) \|^2]  \ ,  \\
	\end{aligned}
\end{equation}
where  $c_1 = \frac{80c_0\eta_{x} L^2}{1-\lambda}\Big ( 1+ \frac{(1-p)}{pb}\Big)$, $c_2 = \frac{80c_0\eta_{x} L^2}{1-\lambda}\Big ( 1+ \frac{(1-p)}{pb}\Big)$, $c_3 = (1-\lambda)\eta_x$, $c_4 = (1-\lambda)c_0\eta_x$, $c_5 =\frac{5\eta_x}{p}$, $c_6 =  \frac{5c_0\eta_x}{p}$. 

%Based on this potential function, we have the following proposition. 
\begin{proposition} \label{proposition_p}
Given Assumptions~\ref{assumption_pl}-\ref{assumption_graph}, by setting $\eta_{x} =\frac{\eta_y}{10c_0}$, $\eta_x \leq \min\{\eta_{x,1}, \eta_{x,2}, \frac{1}{L_g} \}$, $\eta_y \leq \min\{\eta_{y,1}, \eta_{y,2}, \frac{1}{L}\}$, where
\begin{equation}
\begin{aligned}
    & \eta_{x,1} = \frac{(1-\lambda)^2\mu}{44L^2\sqrt{ 1+ (1-p)/(pb)}} \ , \\
    & \eta_{x,2} =  \frac{\mu}{L^2}\frac{-6+\sqrt{36+ 12800 (1  +  (1-p)/(pb))}}{12800  (1  + (1-p)/(pb))} \ , \\
    & \eta_{y,1} = \frac{(1-\lambda)^2}{11L\sqrt{ 1+ (1-p)/(pb)}} \ , \\
    & \eta_{y,2} =\frac{1}{L}\frac{-\frac{1}{2} + \sqrt{\frac{1}{4}+80(1 +  (1-p)/(pb))}}{80(1 +  (1-p)/(pb))} \ , \\
\end{aligned}
\end{equation}
Algorithm~\ref{dpage} satisfies
\begin{equation}
    \begin{aligned}
        & \quad \mathbb{E}[g(\bar{\tilde{x}}_{s+1})-g(x_*)] + \frac{c_0 \eta_{x}}{\eta_y}\mathbb{E}[g(\bar{\tilde{x}}_{s+1})-f(\bar{\tilde{x}}_{s+1}, \bar{\tilde{y}}_{s+1})]\\
        & \leq  \frac{2}{\eta_x \mu R} \Big(\mathbb{E}[g(\bar{\tilde{x}}_{s})-g(x_*)] + \frac{c_0 \eta_{x}}{\eta_y}\mathbb{E}[g(\bar{\tilde{x}}_{s})-f(\bar{\tilde{x}}_{s}, \bar{\tilde{y}}_{s})] \Big)\\
        & \quad + \frac{2}{\eta_x \mu R}\hat{\mathcal{P}}_{s, 0} \ , 
    \end{aligned}
\end{equation}
where $\hat{\mathcal{P}}_{s, 0}$ denotes the last six terms of $\mathcal{P}_{s, 0}$.
\end{proposition}

\begin{remark}
In terms of Proposition~\ref{proposition_p}, to achieve the linear convergence rate, $\hat{\mathcal{P}}_{s, 0}$ should be zero. 
In terms of Algorithm~\ref{dpage}, the full gradient is computed in the first iteration so that the last four terms of the potential function $\hat{\mathcal{P}}_{s, 0}$ are zero values. Then,  we can use the multi-round communication to let $\mathbb{E}[\|X_{s, 0} - \bar{X}_{s, 0}\|_F^2]=0$ and $\mathbb{E}[\|Y_{s, 0} - \bar{Y}_{s, 0}\|_F^2]=0$ to make Algorithm~\ref{dpage}  achieve  the linear convergence rate. That is the reason why we should use the multi-round communication stage-wisely. Otherwise, the consensus errors $\mathbb{E}[\|X_{s, 0} - \bar{X}_{s, 0}\|_F^2]$ and $\mathbb{E}[\|Y_{s, 0} - \bar{Y}_{s, 0}\|_F^2]$ prevents Algorithm~\ref{dpage} from achieving  linear convergence rate.
\end{remark}

\begin{theorem} \label{theorem_p}
Given Assumptions~\ref{assumption_pl}-\ref{assumption_graph}, by setting $\eta_x=O(\frac{(1-\lambda)^2}{\kappa^2L\sqrt{ 1+ (1-p)/(pb)}})$, $\eta_y=O(\frac{(1-\lambda)^2}{L\sqrt{ 1+ (1-p)/(pb)}})$,  $R= \frac{4}{\eta_x\mu }$, Algorithm~\ref{dpage} satisfies
\begin{equation}
    \begin{aligned}
        &  \mathcal{M}_{s+1}   \leq \frac{1}{2} \mathcal{M}_s \ ,
    \end{aligned}
\end{equation}
which implies the linear convergence rate. 
\end{theorem}

\begin{corollary} \label{corollary_p}
In terms of  Theorem~\ref{theorem_p}, to find the solution such that $\mathbb{E}[g(\bar{\tilde{x}}_{s})-g(x_*)] \leq \epsilon$ and $\mathbb{E}[g(\bar{\tilde{x}}_{s})-f(\bar{\tilde{x}}_{s}, \bar{\tilde{y}}_{s})]\leq 10\epsilon$, by setting $b=\sqrt{n}$ and $p=\frac{b}{n+b}$, we can get $\eta_x=O(\frac{(1-\lambda)^2}{\kappa^2 L})$, $\eta_y=O(\frac{(1-\lambda)^2}{L})$,  $R= O(\frac{\kappa^3}{(1-\lambda)^2})$, the iteration complexity is $O(\frac{\kappa^3}{(1-\lambda)^2}\log\frac{1}{\epsilon})$, the sample complexity is $O((n+\frac{ \sqrt{n}\kappa^3}{(1-\lambda)^2})\log\frac{1}{\epsilon})$, and the communication complexity is $ O((D+\frac{\kappa^3}{(1-\lambda)^2})\log\frac{1}{\epsilon})$. 

\end{corollary}

\begin{remark}
    When the communication graph is fully connected, i.e., $1-\lambda=1$, our sample complexity can match the state-of-the-art algorithm SPIDER-GDA \cite{chenfaster} under the single-machine setting. 
\end{remark}

To the best of our knowledge, this is the first decentralized optimization algorithm that can achieve the linear convergence rate for nonconvex-nonconcave optimization problems satisfying PL conditions. 

\subsection{Convergence Rate of Algorithm~\ref{alg_dsvrgd_z}}
To establish the convergence rate of Algorithm~\ref{alg_dsvrgd_z}, we introduce the following potential function:
\begin{equation}
	\begin{aligned}
		& \mathcal{P}_{s, r}  = \mathbb{E}[g(\bar{x}_{s, r})-g(x_*)] + \frac{c_0 \eta_{x}}{\eta_y}\mathbb{E}[g(\bar{x}_{s, r})-f(\bar{x}_{s, r}, \bar{y}_{s, r})] \\
  & \quad + c_1 \frac{1}{K}\mathbb{E}[\|X_{s, r} - \bar{X}_{s, r}\|_F^2] + c_2  \frac{1}{K}\mathbb{E}[\|Y_{s, r} - \bar{Y}_{s, r}\|_F^2] \\
  & \quad + c_3  \frac{1}{K}\mathbb{E}[\| U_{s, r} - \bar{U}_{s, r} \|_F^2] + c_4  \frac{1}{K} \mathbb{E}[\| V_{s, r} - \bar{V}_{s, r} \|_F^2] \\
		& \quad + c_5 \frac{1}{K}\sum_{k=1}^{K}\mathbb{E}[\|\phi^{(k)}(x^{(k)}_{s,r}, y^{(k)}_{s,r})  -\nabla_x f^{(k)}(x^{(k)}_{s,r}, y^{(k)}_{s,r}) \|^2]  \\
		& \quad + c_6 \frac{1}{K}\sum_{k=1}^{K}\mathbb{E}[\|\psi^{(k)}(x^{(k)}_{s,r}, y^{(k)}_{s,r})  -\nabla_y f^{(k)}(x^{(k)}_{s,r}, y^{(k)}_{s,r}) \|^2]    \\
		& \quad + c_7\frac{1}{K}\sum_{k=1}^{K}\mathbb{E}[\frac{1}{n} \sum_{j=1}^{n}\|\nabla_{{x}} f^{(k)}_{j}({x}_{s, r}^{(k)}, {y}_{s, r}^{(k)})-{g}_{s, r}^{(k)}(j)\|^{2}]  \\
		& \quad + 	c_8\frac{1}{K}\sum_{k=1}^{K}\mathbb{E}[\frac{1}{n} \sum_{j=1}^{n}\|\nabla_{{y}} f^{(k)}_{j}({x}_{s, r}^{(k)}, {y}_{s, r}^{(k)})-{h}_{s, r}^{(k)}(j)\|^{2}]  \ ,  \\
	\end{aligned}
\end{equation}
where $c_1=c_2= \frac{c_0\eta_xL^2}{(1- \lambda)} \Big(52 +  \frac{160 }{\rho b_{1}}  +  \frac{1920\rho n^2 }{b_{1}^3}  \Big)$, $c_3 = (1-\lambda) \eta_x$, $c_4 =  (1-\lambda) c_0\eta_x$, $c_5 = \frac{5\eta_x}{\rho}$, $c_6 = \frac{5 c_0\eta_x }{\rho}$, $c_7 =  \frac{30n\rho\eta_x}{b_{1}^2}$, $c_8 =  \frac{30n\rho c_0\eta_x}{b_{1}^2}$.
Then, we can obtain the following proposition. 
\begin{proposition} \label{proposition_z}
Given Assumptions~\ref{assumption_pl}-\ref{assumption_graph}, by setting $\eta_{x} =\frac{\eta_y}{10c_0}$, $\eta_x \leq \min\{\eta_{x,1}, \eta_{x,2}, \frac{1}{L_g} \}$, $\eta_y \leq \min\{\eta_{y,1}, \eta_{y,2}, \frac{1}{L}\}$, where
\begin{equation}
\begin{aligned}
    & \eta_{x,1} = \frac{(1- \lambda)^2\mu}{4L^2\sqrt{76 +  240/(\rho b_{1})  + 1880\rho n^2/{b_{1}^3}  } } \ , \\
    & \eta_{x,2} = \frac{\mu}{L^2}\frac{-6+\sqrt{36 + 80( 6 +{20 }/{(\rho b_{1})}  +  {240\rho n^2 }/{b_{1}^3} )}}{1280 ( 6 +{20 }/{(\rho b_{1})}  +  {240\rho n^2 }/{b_{1}^3} ) } \ , \\
    & \eta_{y,1} =\frac{(1- \lambda)^2}{L\sqrt{76 +  240/(\rho b_{1})  + 1880\rho n^2/{b_{1}^3}  }} \ , \\
    & \eta_{y,2} =\frac{1}{L}\frac{-{1}/{2} +\sqrt{{1}/{4}+8 ( 6 +{20 }/{(\rho b_{1})}  +  {240\rho n^2 }/{b_{1}^3} )}}{8 ( 6 +{20 }/{(\rho b_{1})}  +  {240\rho n^2 }/{b_{1}^3} )} \ , \\
\end{aligned}
\end{equation}
Algorithm~\ref{alg_dsvrgd_z} satisfies
\begin{equation}
    \begin{aligned}
        & \quad \mathbb{E}[g(\bar{\tilde{x}}_{s+1})-g(x_*)] + \frac{c_0 \eta_{x}}{\eta_y}\mathbb{E}[g(\bar{\tilde{x}}_{s+1})-f(\bar{\tilde{x}}_{s+1}, \bar{\tilde{y}}_{s+1})] \\
        & \leq  \frac{2}{\eta_x \mu R} \Big(\mathbb{E}[g(\bar{\tilde{x}}_{s})-g(x_*)] + \frac{c_0 \eta_{x}}{\eta_y}\mathbb{E}[g(\bar{\tilde{x}}_{s})-f(\bar{\tilde{x}}_{s}, \bar{\tilde{y}}_{s})] \Big) \\
		& \quad   +  O(\frac{2}{ \mu R}) \frac{1}{K}(\mathbb{E}[\|X_{s, 0} - \bar{X}_{s, 0}\|_F^2]  + \mathbb{E}[\|Y_{s, 0} - \bar{Y}_{s, 0}\|_F^2] )\\
		& \quad  +O(\frac{c_0}{\mu R} )\frac{n-b_{0}}{(n-1)b_{0}}\frac{1}{K}\sum_{k=1}^{K}\frac{1}{n}\sum_{i=1}^{n} \mathbb{E}[\|\nabla_{{x}} f^{(k)}_i({x}_{s, 0}^{(k)}, {y}_{s, 0}^{(k)})\|^2]   \\
		& \quad  +O(\frac{c_0}{\mu R})\frac{n-b_{0}}{(n-1)b_{0}}\frac{1}{K}\sum_{k=1}^{K}\frac{1}{n}\sum_{i=1}^{n} \mathbb{E}[\|\nabla_{{y}} f^{(k)}_i({x}_{s, 0}^{(k)}, {y}_{s, 0}^{(k)})\|^2 ]\\
		& \quad +O(\frac{2c_0}{\mu R})(1-\frac{b_{0}}{n}) \frac{1}{K}\sum_{k=1}^{K}\frac{1}{n}\sum_{j=1}^{n}\mathbb{E}[\|\nabla_{{x}} f^{(k)}_{j}({x}_{s, 0}^{(k)}, {y}_{s, 0}^{(k)})\|^{2}]  \\
		& \quad + O(\frac{2c_0}{\mu R})(1-\frac{b_{0}}{n}) \frac{1}{K}\sum_{k=1}^{K}\frac{1}{n}\sum_{j=1}^{n}\mathbb{E}[\|\nabla_{{y}} f^{(k)}_{j}({x}_{s, 0}^{(k)}, {y}_{s, 0}^{(k)})\|^{2}]  \ . \\
    \end{aligned}
\end{equation}

% \begin{equation}
%     \begin{aligned}
%         & \quad \mathbb{E}[g(\bar{\tilde{x}}_{s+1})-g(x_*)] + \frac{c_0 \eta_{x}}{\eta_y}\mathbb{E}[g(\bar{\tilde{x}}_{s+1})-f(\bar{\tilde{x}}_{s+1}, \bar{\tilde{y}}_{s+1})]\\
%         & \leq \frac{2\mathcal{P}_{s, 0}}{\mu\eta_x R} \ . 
%     \end{aligned}
% \end{equation}
\end{proposition}

Similarly, we can get the linear convergence rate for Algorithm~\ref{alg_dsvrgd_z} as follows. 
\begin{theorem} \label{theorem_z}
Given Assumptions~\ref{assumption_pl}-\ref{assumption_graph}, by setting $b_0=n$, $b_1=\sqrt{n}$, $\rho=\frac{b_1}{2n}$,  $\eta_x=O(\frac{(1-\lambda)^2}{\kappa^2L})$, $\eta_y=O(\frac{(1-\lambda)^2}{L})$,  $R= \frac{4}{\eta_x\mu }$, Algorithm~\ref{alg_dsvrgd_z} satisfies
\begin{equation}
    \begin{aligned}
        &  \mathcal{M}_{s+1}   \leq \frac{1}{2} \mathcal{M}_s \ ,
    \end{aligned}
\end{equation}
which implies the linear convergence rate. 
\end{theorem}

\begin{corollary} \label{corollary_z}
In terms of  Theorem~\ref{theorem_z}, to find the solution such that $\mathbb{E}[g(\bar{\tilde{x}}_{s})-g(x_*)] \leq \epsilon$ and $\mathbb{E}[g(\bar{\tilde{x}}_{s})-f(\bar{\tilde{x}}_{s}, \bar{\tilde{y}}_{s})]\leq 10\epsilon$, by setting the hyperparameters as shown in Theorem~\ref{theorem_z}, the iteration complexity of Algorithm~\ref{alg_dsvrgd_z} is $O(\frac{\kappa^3}{(1-\lambda)^2}\log\frac{1}{\epsilon})$, the sample complexity is $O((n+\frac{ \sqrt{n}\kappa^3}{(1-\lambda)^2})\log\frac{1}{\epsilon})$, and the communication complexity is $ O((D+\frac{\kappa^3}{(1-\lambda)^2})\log\frac{1}{\epsilon})$. 

\end{corollary}

%\paragraph{Proof Sketch.}
%To prove Proposition~\ref{proposition_p} and Proposition~\ref{proposition_z}, we should bound each term in the proposed potential functions. However, it is challenging. For instance, when bounding $\mathbb{E}[g(\bar{x}_{s, r})-f(\bar{x}_{s, r}, \bar{y}_{s, r})]$, existing works \cite{yang2020global,chenfaster} under the single-machine setting either focus on unbiased gradient estimators or provide the upper bound based on the variance-reduced gradient estimator, i.e., $\psi^{(k)}(x^{(k)}_{s,r}, y^{(k)}_{s,r})$, rather than the original gradient $\nabla_y f(\bar{x}_{s, r}, \bar{y}_{s, r})$. As such, existing theoretical analysis strategies cannot be applied to our algorithms. Moreover, the decentralized communication also brings new challenges to bound $\mathbb{E}[g(\bar{x}_{s, r})-f(\bar{x}_{s, r}, \bar{y}_{s, r})]$. To address these challenges, we develop a new analysis strategy and then provide the desired bound for  $\mathbb{E}[g(\bar{x}_{s, r})-f(\bar{x}_{s, r}, \bar{y}_{s, r})]$ in Lemma~2, which facilitates the proof of our algorithms. We believe our strategy can also be applied to the single-machine setting to address the challenges caused by the biased gradient estimator. Hence, our techniques are novel and have important contributions to this class of optimization problems.

\section{Experiment}
In this section, we conduct extensive experiments to verify the performance of our algorithms. 

\subsection{Experimental Setup}

\paragraph{Synthetic Data.} To verify the performance of our algorithms, we first focus on the following two player Polyak-Łojasiewicz game:
\begin{equation} \label{eq_ex_1_loss}
    \min_{{x}\in \mathbb{R}^{10}} \max_{{y}\in \mathbb{R}^{10}}  \frac{1}{K}\sum_{k=1}^{K} \Big(\frac{1}{2} x^TA^{(k)}x -\frac{1}{2} y^TB^{(k)}y + x^TC^{(k)}y\Big) \ , 
\end{equation}
where $A^{(k)}=\frac{1}{n}\sum_{i=1}^{n}a_i^{(k)}(a_i^{(k)})^{T}$, $B^{(k)}=\frac{1}{n}\sum_{i=1}^{n}b_i^{(k)}(b_i^{(k)})^{T}$, and $C^{(k)}=\frac{1}{n}\sum_{i=1}^{n}c_i^{(k)}(c_i^{(k)})^{T}$. Following \cite{chenfaster}, $a_i^{(k)}$, $b_i^{(k)}$, and $c_i^{(k)}$ are independently drawn from Gaussian distributions $\mathcal{G}(0, \Sigma_{A^{(k)}})$, $\mathcal{G}(0, \Sigma_{B^{(k)}})$, and $\mathcal{G}(0, \Sigma_{C^{(k)}})$, respectively. Here, the covariance matrices $\Sigma_{A^{(k)}}$ and $\Sigma_{B^{(k)}}$ are constructed via $P\Lambda P^T$ \footnote{We ignore the superscript for simplicity.}, where $P\in \mathbb{R}^{10\times 5}$ has orthogonal columns and $\Lambda \in \mathbb{R}^{5\times 5}$ is a diagonal matrix. Additionally, the diagonal elements of $\Lambda$ are uniformly drawn from $[10^{-5}, 1]$. As for $\Sigma_{C^{(k)}}$, it is constructed via $0.1QQ^T$, where $Q\in \mathbb{R}^{10\times 10}$ and the elements of $P$ are drawn from Gaussian distribution $\mathcal{G}(0, 1)$. In this way, the objective function satisfies the PL condition, but not strongly-convex-strongly-concave condition \cite{karimi2016linear,chenfaster}. In our experiment, we set $n=600$.

\paragraph{Real-world Data.} We also verify the performance of our algorithms on  real-world data. Specifically, we focus on the AUC maximization problem for the binary classification task, which is formulated as the following minimax optimization problem: 
\vspace{-10pt}
\begin{equation} \label{eq_ex_2_loss}
	\begin{aligned}
		& \min_{{x}, \hat{x}_1, \hat{x}_2}\max_{y} \frac{1}{K}\sum_{k=1}^{K} \frac{1}{n}\sum_{i=1}^{n}\Big((1-p) ({x}^Ta_i^{(k)} - \hat{x}_1)^2\mathbb{I}_{[b_i^{(k)}=1]} \\
		& + 2(1+y)(p{x}^T{a}_i^{(k)}\mathbb{I}_{[b_i^{(k)}=-1]} - (1-p){x}^T{a}_i^{(k)}\mathbb{I}_{[b_i^{(k)}=1]}) \\
		& + p({x}^Ta_i^{(k)} -\hat{x}_2)^2\mathbb{I}_{[b_i^{(k)}=-1]} - p(1-p)y^2  + \gamma \mathcal{R}({x}) \Big) \ , 
	\end{aligned}
\end{equation}
where ${x}\in \mathbb{R}^d$ is the classifier's parameter, $\hat{x}_1\in \mathbb{R}$  $\hat{x}_2\in \mathbb{R}$, ${y}\in \mathbb{R}$  are the parameters to compute the AUC loss, $(a_i^{(k)}, b_i^{(k)})$ is the $i$-th sample's feature and label on the $k$-th worker, $p$ is the prior probability of positive class, $\mathbb{I}$ is an indicator function, $\mathcal{R}({x})=\sum_{j=1}^{d}\frac{{x}_j^2}{1+{x}_j^2}$ is the regularization term, $\gamma>0$ denotes the hyperparameter.  For this task, we employ three benchmark datasets: a9a, w8a, and ijcnn1, which can be found from LIBSVM Data website \footnote{\url{https://www.csie.ntu.edu.tw/~cjlin/libsvmtools/datasets/}}. In our experiments, we randomly select $20\%$ of samples as the testing set and use the other samples as the training set. 

% $\sum_{j=1}^{d}\frac{\bm{\theta}_j^2}{1+\bm{\theta}_j^2}$
\begin{figure}[h]
 \centering 
% \hspace{-15pt}
 \subfigure[Distance to saddle point]{
  \includegraphics[scale=0.5]{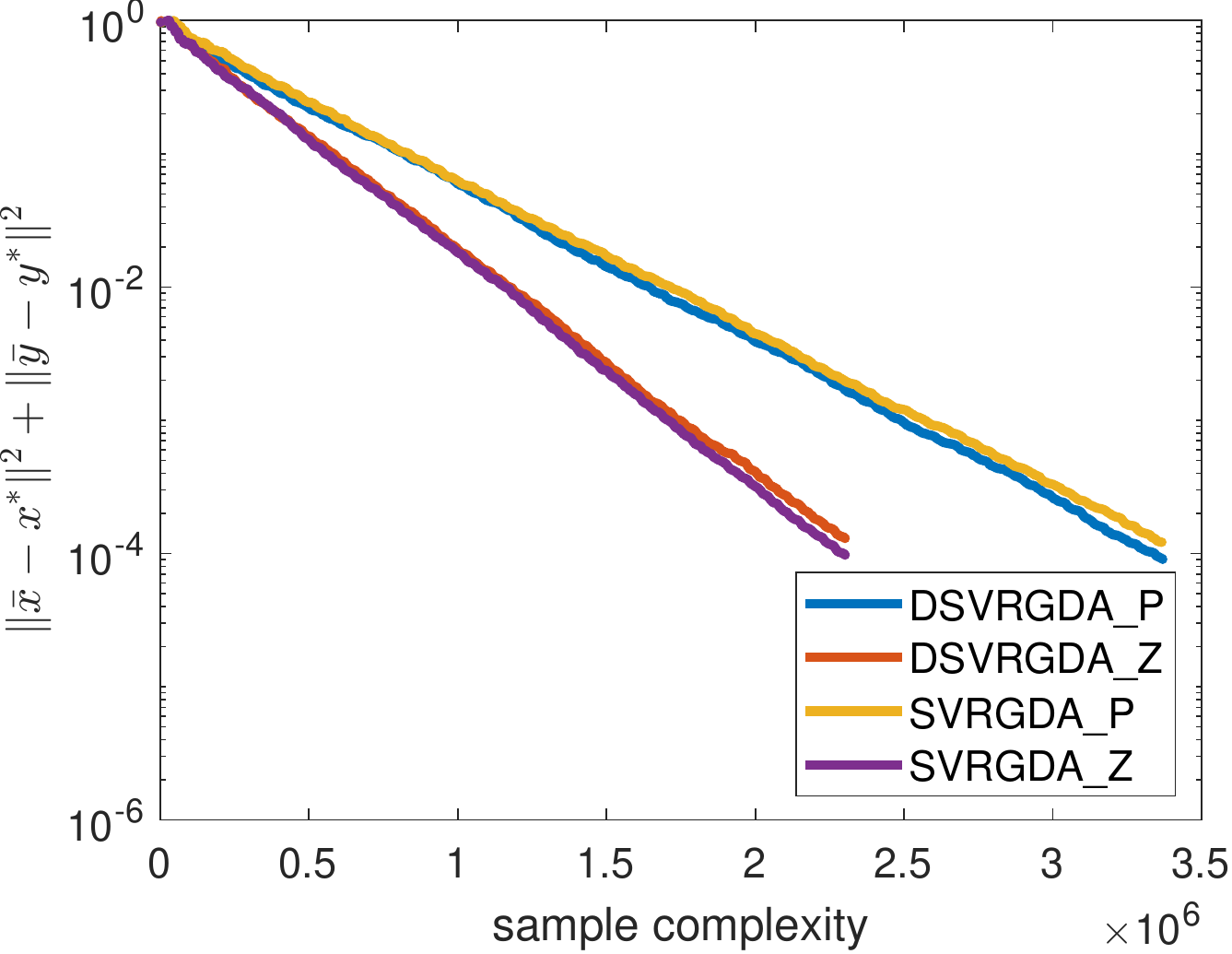}
  \label{fig:synthetic_data_distance}
 }
%  \hspace{-10pt}
   \subfigure[Norm of gradient]{
  \includegraphics[scale=0.5]{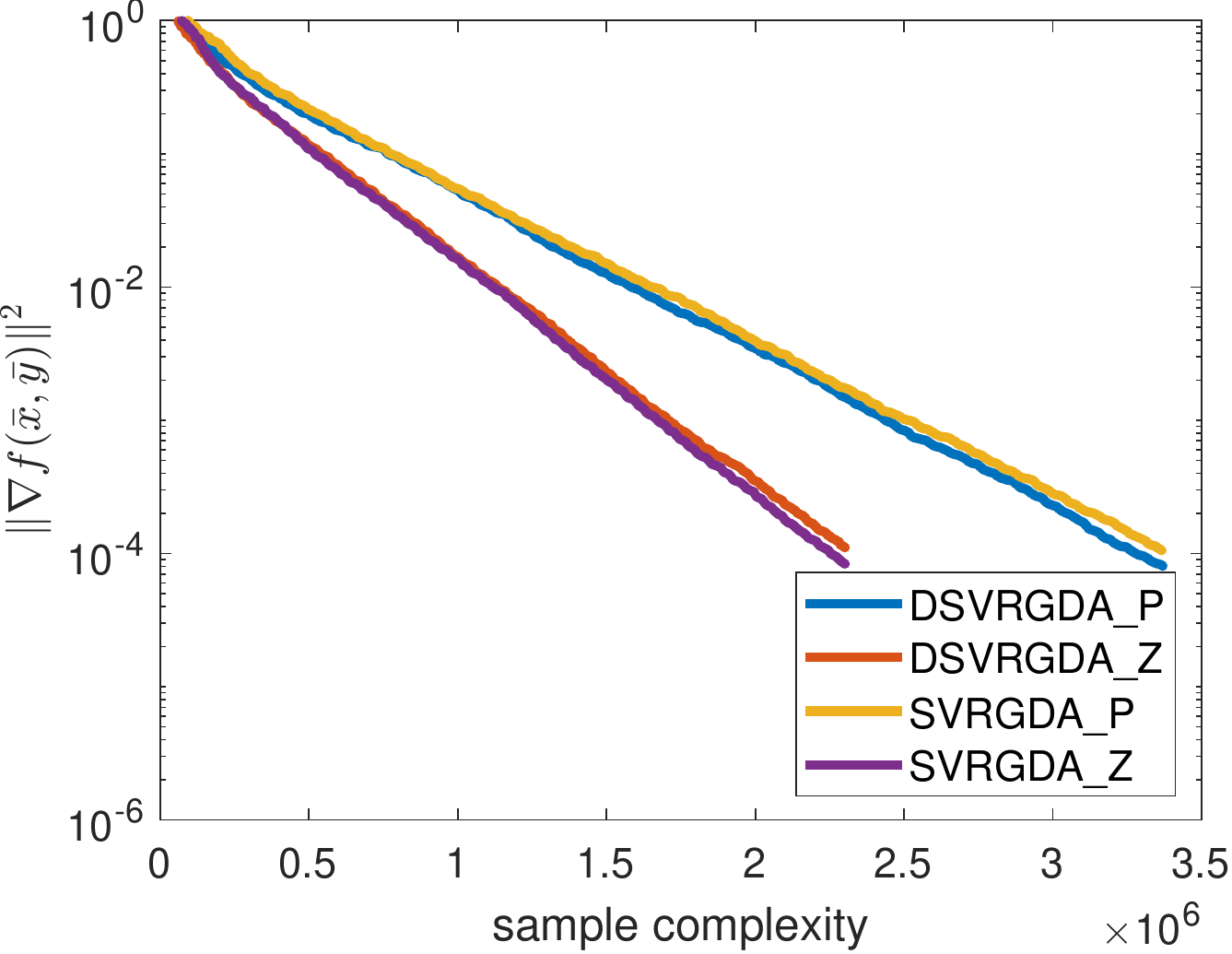}
  \label{fig:synthetic_data_grad_norm}
  }
 \caption{The comparison between different algorithms for Eq.~(\ref{eq_ex_1_loss}). Line graph is used. }
 \label{fig:synthetic_data}
\end{figure}

\paragraph{Experimental Settings. }
Since our algorithms are the first one for the minimax problem satisfying PL conditions, there does not exit baseline algorithms. Therefore, we compare our two algorithms with the single-machine variant: SVRGD-P and SVRGD-Z, where SVRGD-P corresponds to DSVRGDA-P and SVRGD-Z corresponds to DSVRGDA-Z. The details of these baseline algorithms can be found in Appendix B. In our experiments, we use 10 workers and the samples are randomly distributed to the workers. Additionally, we use a line graph to connect these workers. For the single-machine baseline algorithms, we also run them on 10 workers but with a fully connected graph. As such, the effective mini-batch size is the same for all algorithms. 

\begin{figure*}[t]
 \centering 
 \hspace{-15pt}
 \subfigure[w8a]{
  \includegraphics[scale=0.3950]{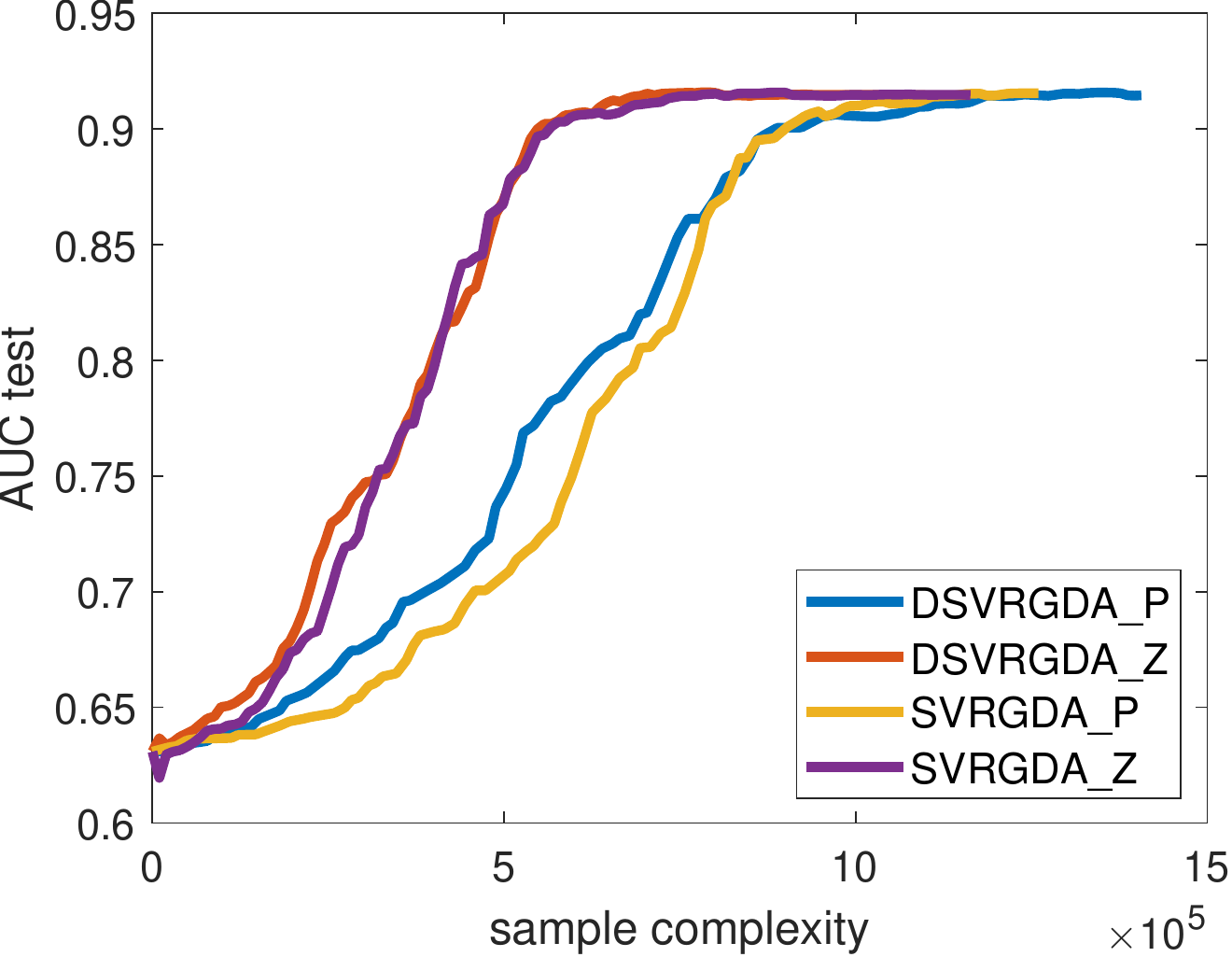}
 }
  \hspace{-1pt}
   \subfigure[a9a]{
  \includegraphics[scale=0.3950]{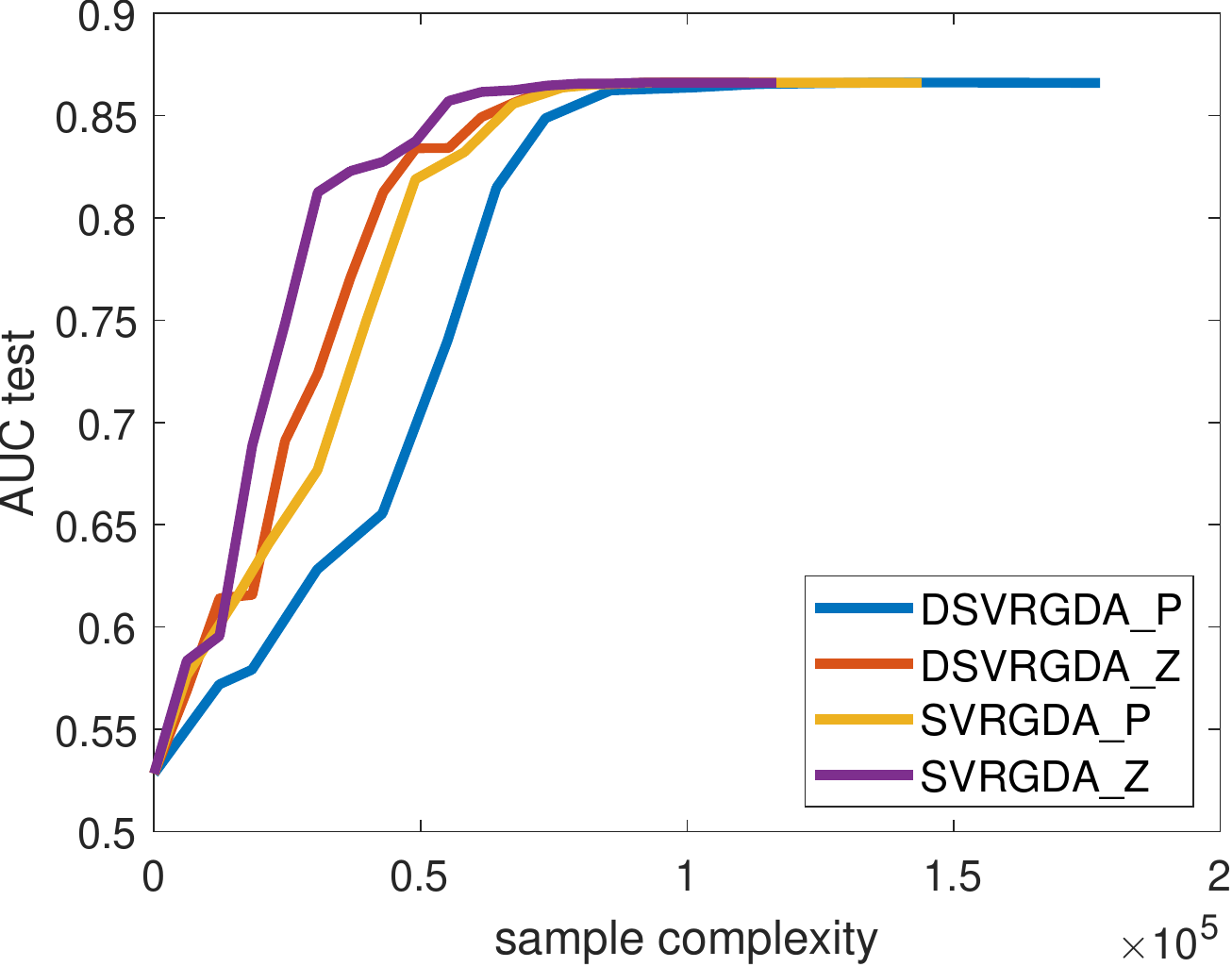}}
  \hspace{-1pt}
    \subfigure[ijcnn1]{
  \includegraphics[scale=0.3950]{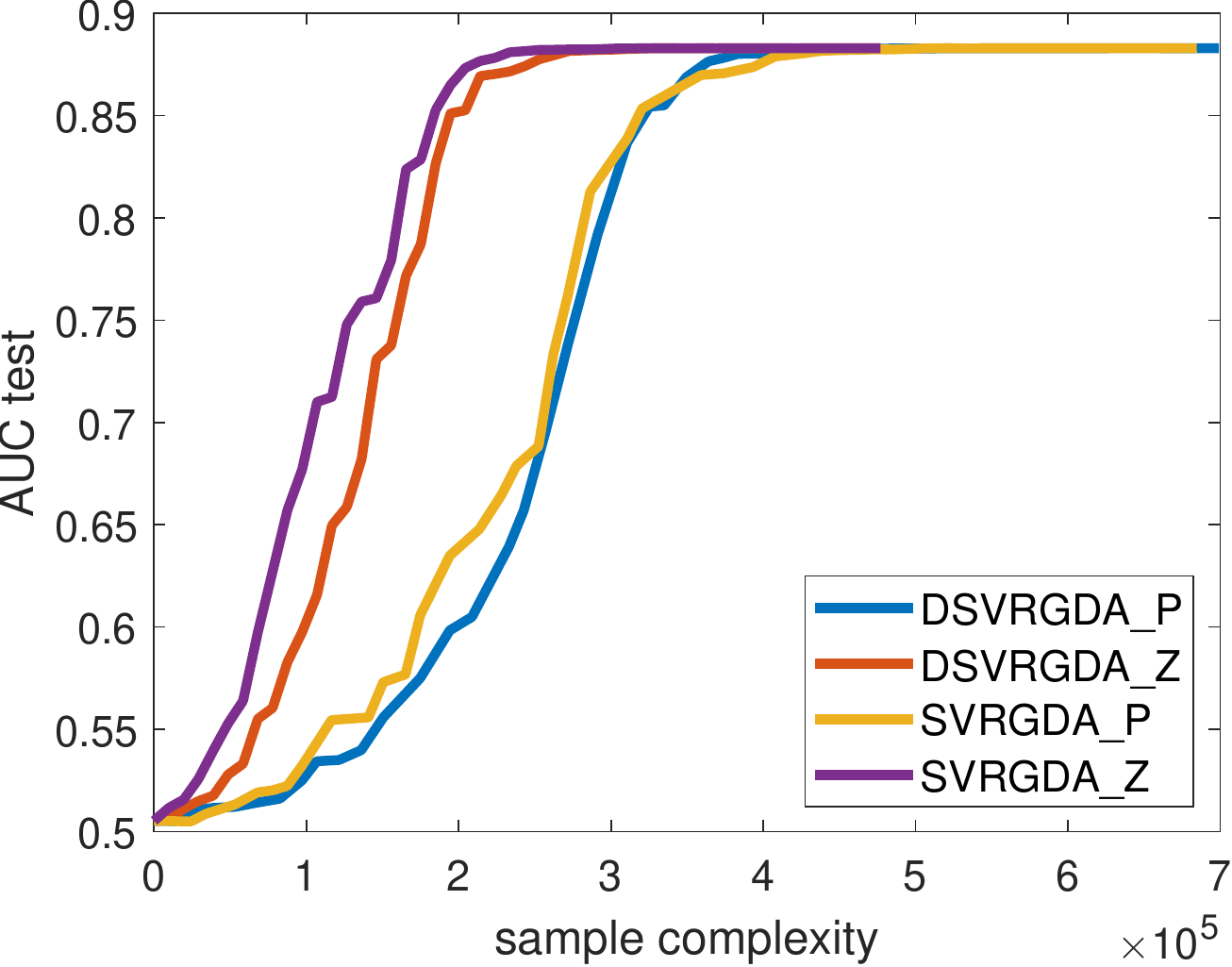}}
 \caption{The comparison between different algorithms for Eq.~(\ref{eq_ex_2_loss}). Line graph is used. }
 \label{fig:real_data_line_graph_nonconvex}
\end{figure*}

\begin{figure*}[!h]
 \centering 
 \hspace{-15pt}
 \subfigure[w8a]{
  \includegraphics[scale=0.390]{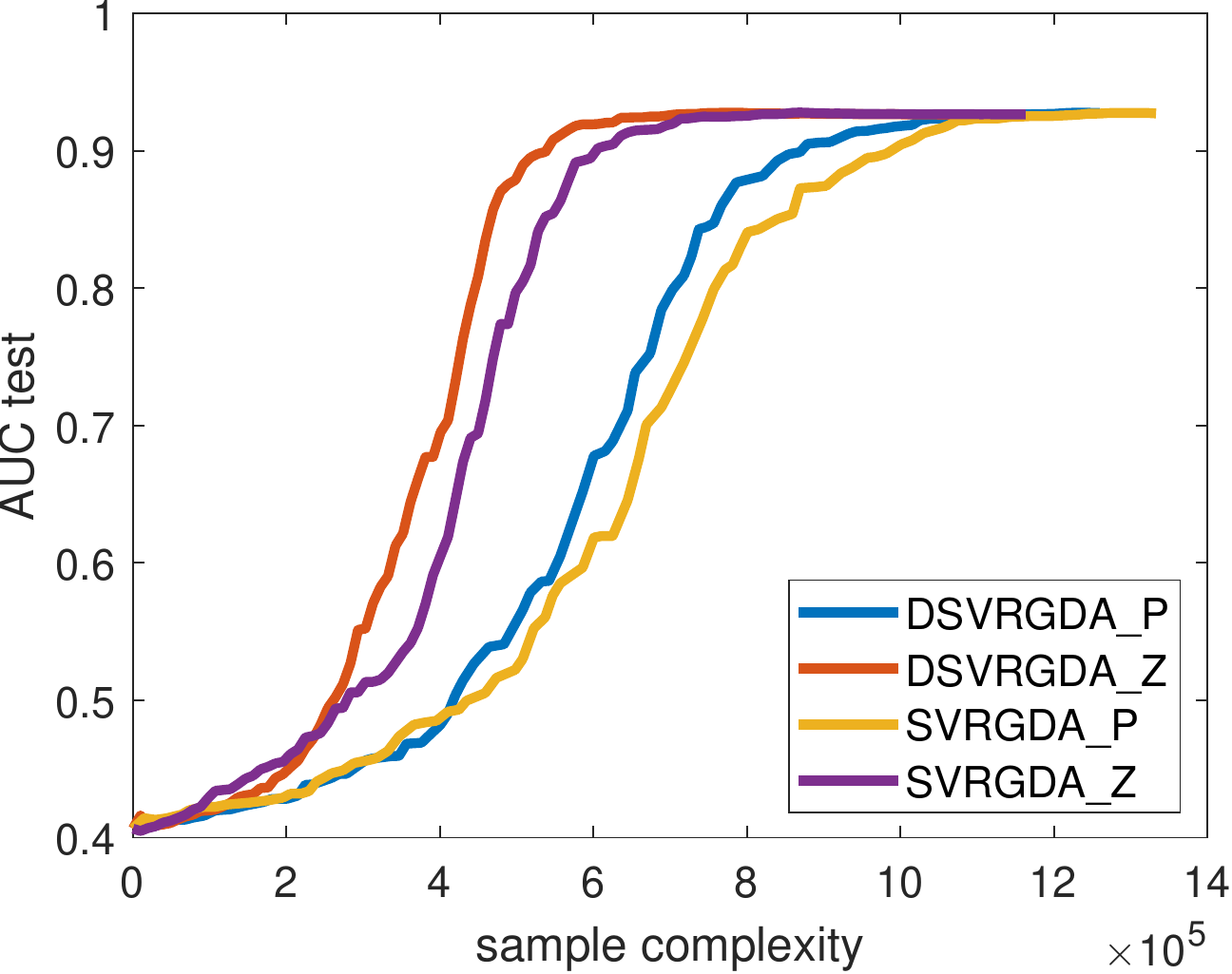}
 }
  \hspace{-1pt}
   \subfigure[a9a]{
  \includegraphics[scale=0.390]{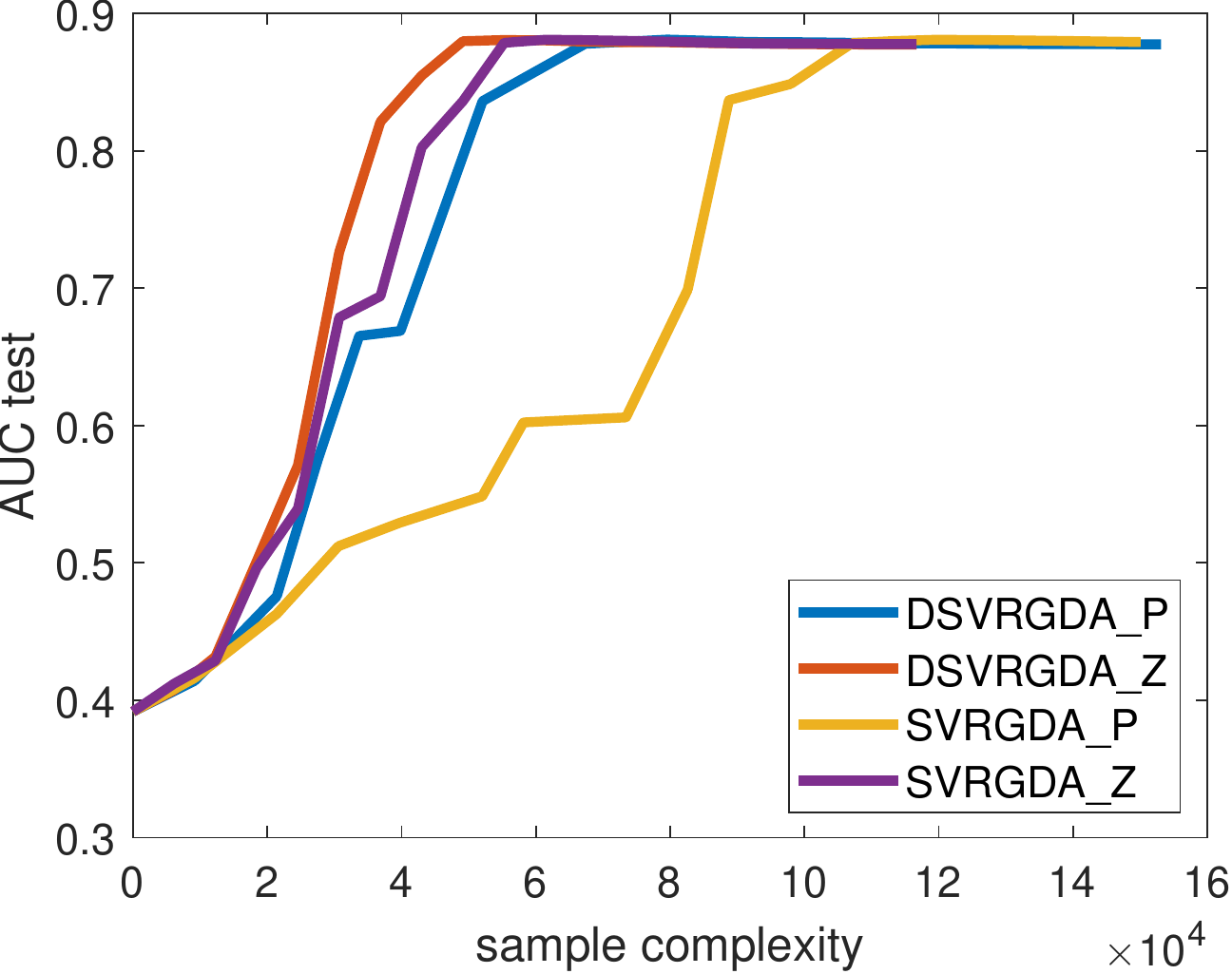}}
  \hspace{-1pt}
    \subfigure[ijcnn1]{
  \includegraphics[scale=0.390]{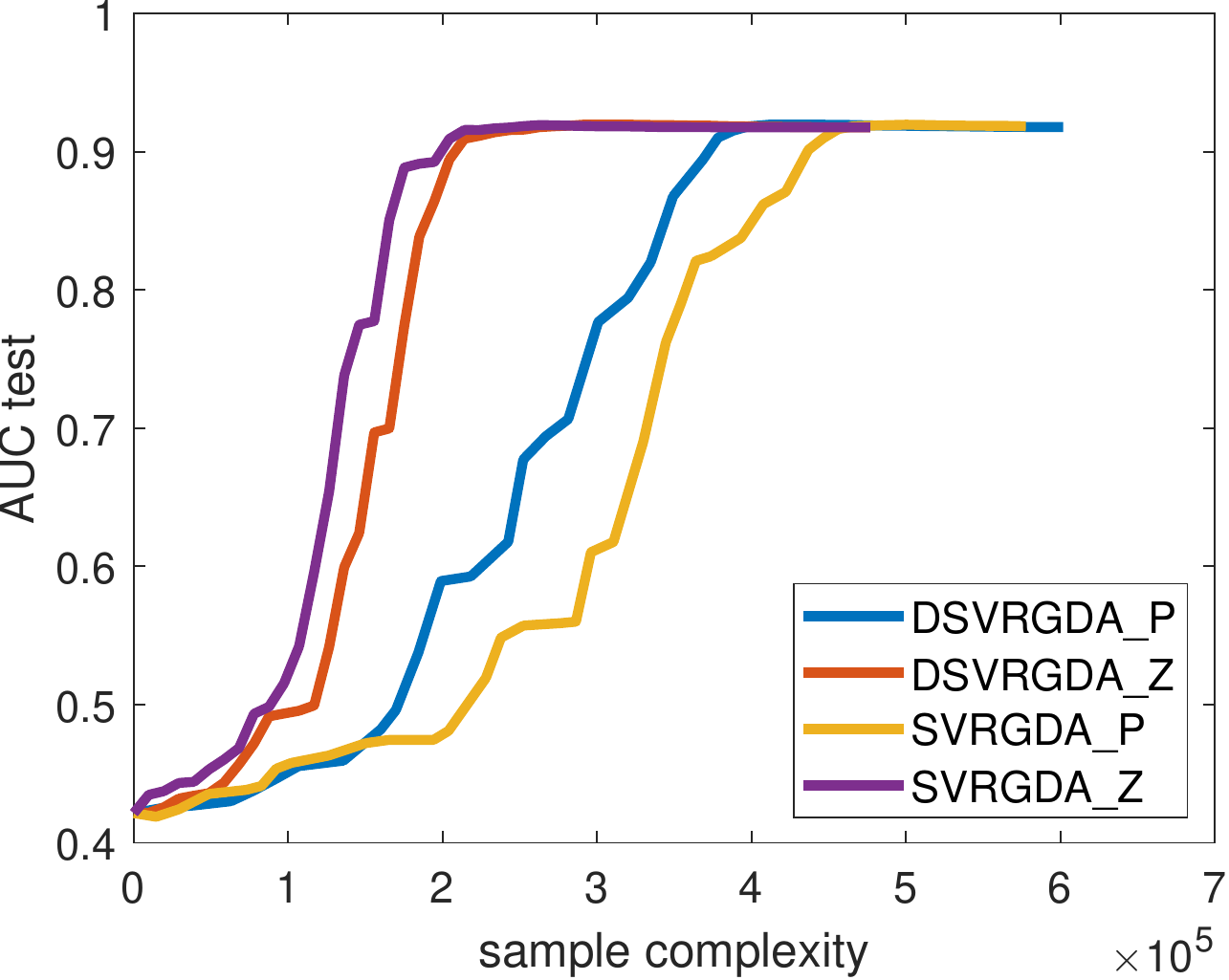}}
 \caption{The comparison between different algorithms for Eq.~(\ref{eq_ex_2_loss}). Random graph is used. }
 \label{fig:real_data_random_graph_nonconvex}
\end{figure*}

In our experiment, the learning rate $\eta_x$ and $\eta_y$ are set to $0.01$ for all algorithms. As for DSVRGDA-P and SVRGD-P, the mini-batch size $b$ is set to $\sqrt{n}$, and the probability $p$ is set to $b/(n+b)$. As for DSVRGDA-Z and SVRGD-Z, the mini-batch size $b_1$ is also set to $\sqrt{n}$, and  $\rho$ is set to $b_1/(2n)$. The number of inner iterations $R$ is set to $\sqrt{n}$. 

% To evaluate the efficiency of DSVRGDA\_P and DSVRGDA\_Z, we compare the AUC performance on test data belong these two algorithms and their corresponding local version SVRGD\_P and SVRGD\_Z. In addition, we conducted experiments on the three benchmark datasets where the regularization term of loss function is l2 norm. 

% We set $\eta_x = 0.01$, $\eta_y = 0.01$ and batch size for each worker = (\#of train data/\#of worker) for all of these algorithms, both of the minibatchsize in each random sampling and the number of inner loop are $\sqrt{bs}$. For decentralized algorithms, we use 10 workers in the experiment. Specifically, we set probabilities in DSVRGDA\_P, SVRGD\_P as 0.1 and $\rho = \sqrt{0.5/bs}$ in DSVRGDA\_Z, SVRGD\_Z.For the synthetic data, we create a dataset with 6000 samples. For the three benchmark datasets: a9a, w8a and ijcnn1, we use the split ratio = 0.2 to get the train data and test data. 

\vspace{-20pt}
\subsection{Experimental Result}
\vspace{-10pt}
In Figure~\ref{fig:synthetic_data}, we show the experimental results on the synthetic data. Specifically, in Figure~\ref{fig:synthetic_data_distance}, we show $\|\bar{x}-x^*\|^2 + \|\bar{y}-y^*\|^2$, which is the distance between the obtained solution from different algorithms  to the optimal solution,  versus the number of sample complexity. We can observe that our two algorithms demonstrate almost the same convergence behavior as the single-machine counterpart, which confirms the correctness of our algorithms. Moreover, the empirical sample complexity of DSVRGDA-Z is smaller than that of DSVRGDA-P. The reason is that DSVRGDA-Z computes the full gradient only in the initial inner iteration. 
In Figure~\ref{fig:synthetic_data_grad_norm}, we plot $\|[\nabla_x f(x, y)^T, \nabla_y f(x, y)^T]^T\|^2$ versus the sample complexity. The observations are similar to Figure~\ref{fig:synthetic_data_distance}. Our decentralized algorithms share the similar convergence behavior with the single-machine counterpart and DSVRGDA-Z enjoys a smaller empirical sample complexity. These observations confirm the correctness and effectiveness of our algorithms. 

In Figure~\ref{fig:real_data_line_graph_nonconvex}, we show the experimental results on the real-world data. In particular, we show the AUC score of the testing set versus the sample complexity. We can still observe that our decentralized algorithms converge to almost the same value with the single-machine counterparts, which confirms the correctness of our algorithms. Moreover, DSVRGDA-Z enjoys a better empirical sample complexity than DSVRDA-P. 

In Figure~\ref{fig:real_data_random_graph_nonconvex}, the communication graph is an  Erdos-Renyi random graph, where the probability for generating edges is $0.5$. Under this setting, we have similar observations as the line graph, which further confirms the effectiveness of our algorithms. The results for Eq.~(\ref{eq_ex_1_loss}) can be found in Appendix A. 

\vspace{-10pt}
\section{Conclusion}
%\vspace{-10pt}
In this paper, we develop two  algorithms for decentralized nonconvex-nonconcave optimization problems that satisfy the PL condition, which demonstrates  how to conduct local updates and perform communication to achieve the linear convergence rate. Moreover, our theoretical analyses provide theoretical evidence for our novel algorithmic design. 
% To the best of our knowledge, this is the first work achieving linear convergence rates for  decentralized nonconvex-nonconcave optimization problems. 
The extensive experimental results confirm the effectiveness of our algorithms.

% \input{supp_exp}

%\newpage
% References
% \bibliographystyle{abbrv}
\bibliography{uai2023-template}

\begin{thebibliography}{30}
\providecommand{\natexlab}[1]{#1}
\providecommand{\url}[1]{\texttt{#1}}
\expandafter\ifx\csname urlstyle\endcsname\relax
  \providecommand{\doi}[1]{doi: #1}\else
  \providecommand{\doi}{doi: \begingroup \urlstyle{rm}\Url}\fi

\bibitem[Chen et~al.(2022{\natexlab{a}})Chen, Yao, and Luo]{chenfaster}
Lesi Chen, Boyuan Yao, and Luo Luo.
\newblock Faster stochastic algorithms for minimax optimization under
  polyak-$\{$$\backslash$L$\}$ ojasiewicz condition.
\newblock In \emph{Advances in Neural Information Processing Systems},
  2022{\natexlab{a}}.

\bibitem[Chen et~al.(2022{\natexlab{b}})Chen, Ye, and Luo]{chen2022simple}
Lesi Chen, Haishan Ye, and Luo Luo.
\newblock A simple and efficient stochastic algorithm for decentralized
  nonconvex-strongly-concave minimax optimization.
\newblock \emph{arXiv preprint arXiv:2212.02387}, 2022{\natexlab{b}}.

\bibitem[Cutkosky and Orabona(2019)]{cutkosky2019momentum}
Ashok Cutkosky and Francesco Orabona.
\newblock Momentum-based variance reduction in non-convex sgd.
\newblock \emph{arXiv preprint arXiv:1905.10018}, 2019.

\bibitem[Fang et~al.(2018)Fang, Li, Lin, and Zhang]{fang2018spider}
Cong Fang, Chris~Junchi Li, Zhouchen Lin, and Tong Zhang.
\newblock Spider: Near-optimal non-convex optimization via stochastic path
  integrated differential estimator.
\newblock \emph{arXiv preprint arXiv:1807.01695}, 2018.

\bibitem[Gao(2022)]{gao2022decentralized}
Hongchang Gao.
\newblock Decentralized stochastic gradient descent ascent for finite-sum
  minimax problems.
\newblock \emph{arXiv preprint arXiv:2212.02724}, 2022.

\bibitem[Goodfellow et~al.(2014{\natexlab{a}})Goodfellow, Pouget-Abadie, Mirza,
  Xu, Warde-Farley, Ozair, Courville, and Bengio]{goodfellow2014generative}
Ian Goodfellow, Jean Pouget-Abadie, Mehdi Mirza, Bing Xu, David Warde-Farley,
  Sherjil Ozair, Aaron Courville, and Yoshua Bengio.
\newblock Generative adversarial nets.
\newblock \emph{Advances in neural information processing systems}, 27,
  2014{\natexlab{a}}.

\bibitem[Goodfellow et~al.(2014{\natexlab{b}})Goodfellow, Shlens, and
  Szegedy]{goodfellow2014explaining}
Ian~J Goodfellow, Jonathon Shlens, and Christian Szegedy.
\newblock Explaining and harnessing adversarial examples.
\newblock \emph{arXiv preprint arXiv:1412.6572}, 2014{\natexlab{b}}.

\bibitem[Huang et~al.(2020)Huang, Gao, Pei, and Huang]{huang2020accelerated}
Feihu Huang, Shangqian Gao, Jian Pei, and Heng Huang.
\newblock Accelerated zeroth-order momentum methods from mini to minimax
  optimization.
\newblock \emph{arXiv e-prints}, pages arXiv--2008, 2020.

\bibitem[Johnson and Zhang(2013)]{johnson2013accelerating}
Rie Johnson and Tong Zhang.
\newblock Accelerating stochastic gradient descent using predictive variance
  reduction.
\newblock \emph{Advances in neural information processing systems}, 26, 2013.

\bibitem[Karimi et~al.(2016)Karimi, Nutini, and Schmidt]{karimi2016linear}
Hamed Karimi, Julie Nutini, and Mark Schmidt.
\newblock Linear convergence of gradient and proximal-gradient methods under
  the polyak-{\l}ojasiewicz condition.
\newblock In \emph{Machine Learning and Knowledge Discovery in Databases:
  European Conference, ECML PKDD 2016, Riva del Garda, Italy, September 19-23,
  2016, Proceedings, Part I 16}, pages 795--811. Springer, 2016.

\bibitem[Li and Richt{\'a}rik(2021)]{li2021zerosarah}
Zhize Li and Peter Richt{\'a}rik.
\newblock Zerosarah: Efficient nonconvex finite-sum optimization with zero full
  gradient computation.
\newblock \emph{arXiv preprint arXiv:2103.01447}, 2021.

\bibitem[Li et~al.(2021)Li, Bao, Zhang, and Richt{\'a}rik]{li2021page}
Zhize Li, Hongyan Bao, Xiangliang Zhang, and Peter Richt{\'a}rik.
\newblock Page: A simple and optimal probabilistic gradient estimator for
  nonconvex optimization.
\newblock In \emph{International conference on machine learning}, pages
  6286--6295. PMLR, 2021.

\bibitem[Lian et~al.(2017)Lian, Zhang, Zhang, Hsieh, Zhang, and
  Liu]{lian2017can}
Xiangru Lian, Ce~Zhang, Huan Zhang, Cho-Jui Hsieh, Wei Zhang, and Ji~Liu.
\newblock Can decentralized algorithms outperform centralized algorithms? a
  case study for decentralized parallel stochastic gradient descent.
\newblock In \emph{Advances in Neural Information Processing Systems}, pages
  5330--5340, 2017.

\bibitem[Lin et~al.(2020)Lin, Jin, and Jordan]{lin2020gradient}
Tianyi Lin, Chi Jin, and Michael Jordan.
\newblock On gradient descent ascent for nonconvex-concave minimax problems.
\newblock In \emph{International Conference on Machine Learning}, pages
  6083--6093. PMLR, 2020.

\bibitem[Liu et~al.(2022)Liu, Zhu, and Belkin]{liu2022loss}
Chaoyue Liu, Libin Zhu, and Mikhail Belkin.
\newblock Loss landscapes and optimization in over-parameterized non-linear
  systems and neural networks.
\newblock \emph{Applied and Computational Harmonic Analysis}, 59:\penalty0
  85--116, 2022.

\bibitem[Liu et~al.(2019)Liu, Yuan, Ying, and Yang]{liu2019stochastic}
Mingrui Liu, Zhuoning Yuan, Yiming Ying, and Tianbao Yang.
\newblock Stochastic auc maximization with deep neural networks.
\newblock \emph{arXiv preprint arXiv:1908.10831}, 2019.

\bibitem[Lu et~al.(2019)Lu, Zhang, Sun, and Hong]{lu2019gnsd}
Songtao Lu, Xinwei Zhang, Haoran Sun, and Mingyi Hong.
\newblock Gnsd: a gradient-tracking based nonconvex stochastic algorithm for
  decentralized optimization.
\newblock In \emph{2019 IEEE Data Science Workshop, DSW 2019}, pages 315--321.
  Institute of Electrical and Electronics Engineers Inc., 2019.

\bibitem[Luo et~al.(2020)Luo, Ye, Huang, and Zhang]{luo2020stochastic}
Luo Luo, Haishan Ye, Zhichao Huang, and Tong Zhang.
\newblock Stochastic recursive gradient descent ascent for stochastic
  nonconvex-strongly-concave minimax problems.
\newblock \emph{arXiv preprint arXiv:2001.03724}, 2020.

\bibitem[Madry et~al.(2017)Madry, Makelov, Schmidt, Tsipras, and
  Vladu]{madry2017towards}
Aleksander Madry, Aleksandar Makelov, Ludwig Schmidt, Dimitris Tsipras, and
  Adrian Vladu.
\newblock Towards deep learning models resistant to adversarial attacks.
\newblock \emph{arXiv preprint arXiv:1706.06083}, 2017.

\bibitem[Nguyen et~al.(2017)Nguyen, Liu, Scheinberg, and
  Tak{\'a}{\v{c}}]{nguyen2017sarah}
Lam~M Nguyen, Jie Liu, Katya Scheinberg, and Martin Tak{\'a}{\v{c}}.
\newblock Sarah: A novel method for machine learning problems using stochastic
  recursive gradient.
\newblock In \emph{International Conference on Machine Learning}, pages
  2613--2621. PMLR, 2017.

\bibitem[Nouiehed et~al.(2019)Nouiehed, Sanjabi, Huang, Lee, and
  Razaviyayn]{nouiehed2019solving}
Maher Nouiehed, Maziar Sanjabi, Tianjian Huang, Jason~D Lee, and Meisam
  Razaviyayn.
\newblock Solving a class of non-convex min-max games using iterative first
  order methods.
\newblock \emph{arXiv preprint arXiv:1902.08297}, 2019.

\bibitem[Scaman et~al.(2017)Scaman, Bach, Bubeck, Lee, and
  Massouli{\'e}]{scaman2017optimal}
Kevin Scaman, Francis Bach, S{\'e}bastien Bubeck, Yin~Tat Lee, and Laurent
  Massouli{\'e}.
\newblock Optimal algorithms for smooth and strongly convex distributed
  optimization in networks.
\newblock In \emph{international conference on machine learning}, pages
  3027--3036. PMLR, 2017.

\bibitem[Sun et~al.(2020)Sun, Lu, and Hong]{sun2020improving}
Haoran Sun, Songtao Lu, and Mingyi Hong.
\newblock Improving the sample and communication complexity for decentralized
  non-convex optimization: Joint gradient estimation and tracking.
\newblock In \emph{International Conference on Machine Learning}, pages
  9217--9228. PMLR, 2020.

\bibitem[Sun et~al.(2018)Sun, Qu, and Wright]{sun2018geometric}
Ju~Sun, Qing Qu, and John Wright.
\newblock A geometric analysis of phase retrieval.
\newblock \emph{Foundations of Computational Mathematics}, 18:\penalty0
  1131--1198, 2018.

\bibitem[Vogels et~al.(2021)Vogels, He, Koloskova, Karimireddy, Lin, Stich, and
  Jaggi]{vogels2021relaysum}
Thijs Vogels, Lie He, Anastasiia Koloskova, Sai~Praneeth Karimireddy, Tao Lin,
  Sebastian~U Stich, and Martin Jaggi.
\newblock Relaysum for decentralized deep learning on heterogeneous data.
\newblock \emph{Advances in Neural Information Processing Systems},
  34:\penalty0 28004--28015, 2021.

\bibitem[Xian et~al.(2021)Xian, Huang, Zhang, and Huang]{xian2021faster}
Wenhan Xian, Feihu Huang, Yanfu Zhang, and Heng Huang.
\newblock A faster decentralized algorithm for nonconvex minimax problems.
\newblock \emph{Advances in Neural Information Processing Systems}, 34, 2021.

\bibitem[Xin et~al.(2020)Xin, Khan, and Kar]{xin2020near}
Ran Xin, Usman~A Khan, and Soummya Kar.
\newblock A near-optimal stochastic gradient method for decentralized
  non-convex finite-sum optimization.
\newblock \emph{arXiv preprint arXiv:2008.07428}, 2020.

\bibitem[Yang et~al.(2020)Yang, Kiyavash, and He]{yang2020global}
Junchi Yang, Negar Kiyavash, and Niao He.
\newblock Global convergence and variance-reduced optimization for a class of
  nonconvex-nonconcave minimax problems.
\newblock \emph{arXiv preprint arXiv:2002.09621}, 2020.

\bibitem[Ying et~al.(2016)Ying, Wen, and Lyu]{ying2016stochastic}
Yiming Ying, Longyin Wen, and Siwei Lyu.
\newblock Stochastic online auc maximization.
\newblock \emph{Advances in neural information processing systems},
  29:\penalty0 451--459, 2016.

\bibitem[Zhang et~al.(2021)Zhang, Liu, Liu, Zhu, and Lu]{zhang2021taming}
Xin Zhang, Zhuqing Liu, Jia Liu, Zhengyuan Zhu, and Songtao Lu.
\newblock Taming communication and sample complexities in decentralized policy
  evaluation for cooperative multi-agent reinforcement learning.
\newblock \emph{Advances in Neural Information Processing Systems},
  34:\penalty0 18825--18838, 2021.

\end{thebibliography}

% \appendix
% \onecolumn
% \input{supp_0}
% \newpage
% \input{supp_1}
% \newpage
% \input{supp_2}
% % \newpage
% \input{supp_3}

\end{document}